%% file: iclr2027_conference.tex
\documentclass{article} 
\usepackage{iclr2027_conference,times}
\usepackage{graphicx}
\usepackage{float}

\input{sections/math_commands.tex}

\definecolor{linkpurple}{RGB}{255,0,0}
\definecolor{SkyBlue}{RGB}{135,206,235}
\definecolor{urlorange}{RGB}{190,90,20}
\usepackage[colorlinks=true, linkcolor=linkpurple, citecolor=SkyBlue, urlcolor=urlorange]{hyperref}
\usepackage{url}
\usepackage{tikz}
\usepackage{booktabs}
\usepackage{array,longtable,needspace,fvextra,tcolorbox,multirow}
\usepackage{pifont}
\usepackage{colortbl}

\definecolor{checkgreen}{RGB}{34,125,78}
\definecolor{crossred}{RGB}{180,75,75}
\definecolor{tableblue}{RGB}{235,246,252}
\newcommand{\rsiYes}{\tikz[baseline=-0.55ex]\node[rounded corners=2pt, fill=checkgreen!14, text=checkgreen, inner xsep=2pt, inner ysep=0.5pt, font=\scriptsize]{\ding{51}};}
\newcommand{\rsiNo}{\tikz[baseline=-0.55ex]\node[rounded corners=2pt, fill=crossred!12, text=crossred, inner xsep=2pt, inner ysep=0.5pt, font=\scriptsize]{\ding{55}};}
\newcounter{finding}
\newcommand{\finding}[2]{\par\refstepcounter{finding}\noindent
  \begin{tikzpicture}\node[draw=SkyBlue!85!black, fill=SkyBlue!8, rounded corners=3pt,
    line width=.45pt, inner xsep=5pt, inner ysep=2pt, text width={\linewidth-11pt}, align=justify,
    font=\footnotesize]{\textbf{Finding~\thefinding: #1} \emph{#2}};\end{tikzpicture}\par}
\newcommand{\plannerline}[2]{#1\,\textcolor{gray}{\fontsize{6}{7}\selectfont (planner:\,\raisebox{-0.15ex}{\includegraphics[height=1.10em]{#2}})}}

\title{MC-EvoBench: Auditing Experience-Driven Self-Improvement in Minecraft Agents}

\author{%
\parbox{\textwidth}{\raggedright\setlength{\baselineskip}{13pt}%
Jingwei~Song$^{1,*}$ \quad
Zhisheng~Chen$^{2,*}$ \quad
Bingfan~Zeng$^{3}$ \quad
Jing~Tang$^{4}$ \quad
Yubo~Hou$^{5}$ \quad
Jinhao~Jing$^{6}$ \quad
Zhengwei~Xie$^{7,\dagger}$ \quad
Linfeng~Zhang$^{1,\dagger}$ \endgraf
\vspace{6pt}
{\normalfont\small
\mbox{$^{1}$Shanghai Jiao Tong University} \quad
\mbox{$^{2}$University of the Chinese Academy of Sciences} \quad
\mbox{$^{3}$South China University of Technology} \quad
\mbox{$^{4}$Huazhong University of Science and Technology} \quad
\mbox{$^{5}$Beihang University} \quad
\mbox{$^{6}$The Chinese University of Hong Kong (Shenzhen)} \quad
\mbox{$^{7}$University of Science and Technology of China} \endgraf}
\vspace{5pt}
{\normalfont\small
\texttt{u3638265@connect.hku.hk}, \texttt{zhisheng.researcher@gmail.com} \endgraf}
}}

\iclrfinalcopy 
\begin{document}

\maketitle

{\renewcommand{\thefootnote}{\fnsymbol{footnote}}
\footnotetext[1]{Equal contribution.}
\footnotetext[2]{Corresponding authors.}}

\vspace{-5mm}

\begin{abstract}
A central challenge in embodied recursive self-improvement is enabling agents to convert interaction history into capabilities that persist beyond one world and transfer to later tasks. A saved memory or successful final action does not by itself establish that a reusable capability was acquired; evaluation must separate capability change from carried-over resources, unequal source exposure, and early execution failures. We construct MC-EvoBench as a controlled Minecraft benchmark with a fixed catalog of 19 registered evolution chains, including 12 primary evaluation chains and seven calibration or probe chains, together with three graph-aligned paired probe families for transfer. The benchmark represents each task as a task-state graph, defines stage-level environment verifiers, assigns matched source and target roles, resets the target world, and permits only a declared agent artifact to cross the source--target boundary. For each probe, no-prior, related, unrelated, and shuffled source conditions are evaluated under matched environment-step budgets and five seeds; we report TSR, WGP, and VMP and retain provenance records. The benchmark reveals a consistent direct-capability gap between controller and planner-backed variants, while related experience yields no stable transfer: it does not improve either tool probe, its small wood-species gain hinges on one seed, and all intervals include zero. Stage audits further show that early wood-collection and crafting bottlenecks often leave later capabilities unobserved. MC-EvoBench thus offers a controlled testbed for durable, transferable embodied capabilities. Code and data are available at \url{https://github.com/xzw-ustc/mc-bench-open-source}.
\end{abstract}

\begin{figure}[H]
    \centering
    \includegraphics[width=0.7\linewidth]{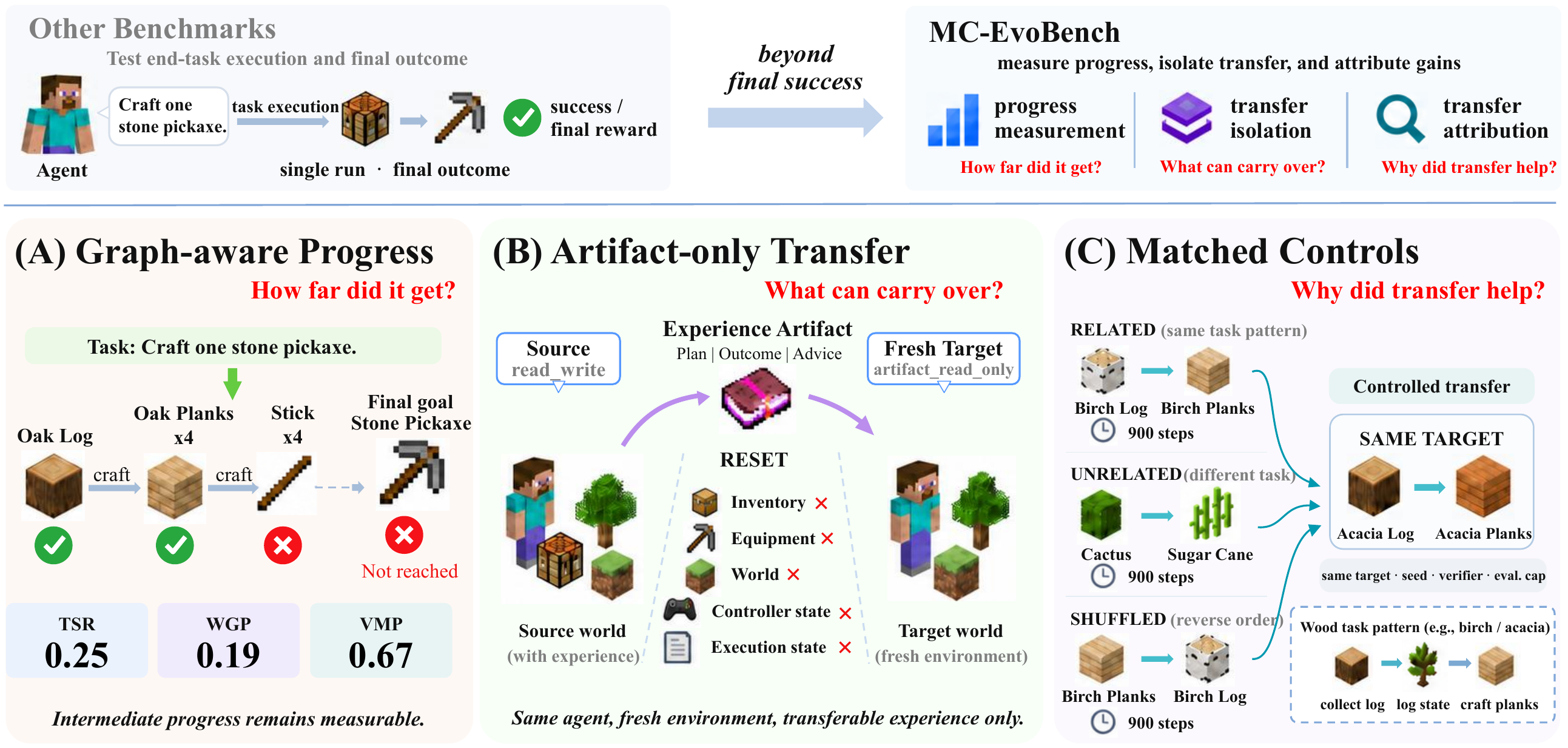}
    \caption{Evaluation mechanism of MC-EvoBench. A conventional benchmark records only whether an agent reaches the goal. MC-EvoBench also measures progress through the task graph, transfers only a declared experience artifact into a reset target world, and compares related, unrelated, and shuffled source conditions. We separate partial progress, physical carryover, and source-specific experience gains from improvements caused by favorable initial states or execution luck.}
    \label{fig:mechanism}
\end{figure}

\input{sections/Introduction.tex}

\input{sections/Related_Work.tex}

\input{sections/MC-EvoBench_Framework_and_Benchmark_Design.tex}

\input{sections/Experimental_Results.tex}

\section{Conclusion}
\label{sec:conclusion}

We introduced MC-EvoBench, a controlled Minecraft benchmark for auditing whether experience yields durable and transferable embodied capabilities. Its task-state graph, stage-level verifiers, and paired reset-target protocol with matched control sources separate capability change from carried-over resources, unequal exposure, and early execution failures. In the reference study, controller-based agents clearly outperform planner-backed variants, yet the reference method gains no stable target benefit from related experience: the only positive contrast, on wood species, is driven by a single seed, and all intervals include zero even though related sources earn more verified progress and, on the longer probes, more computation. Most runs also end before crafting sticks, leaving later capabilities unobserved. These conclusions are limited to the tested agents, probes, seeds, and budgets; retention, composition, stronger agents, and more seeds are natural next steps.

\newpage
\subsection*{AI use statement}

We used generative AI tools for English translation, language editing,
shortening text, and LaTeX formatting. We did not use AI to generate data,
run experiments, or draw the final conclusions. The authors checked all
citations, equations, experimental settings, and results. They reviewed all
AI-assisted text and take full responsibility for the paper.

\subsection*{Ethics statement}

This work evaluates agents in a virtual Minecraft environment. It does not
involve human participants, personal data, or physical-world experiments.
We use environment states and agent interaction traces to measure task
execution and experience transfer. Any released code, task definitions,
assets, and logs will follow their applicable licenses and include
documentation for reproduction. The results apply to the tested methods,
tasks, environments, and budgets. They do not establish that these agents
are safe or capable in real-world settings.

\subsection*{Reproducibility statement}

Section~\ref{sec:framework} defines the task-state graph, source--target
protocol, and metrics. Section~\ref{sec:results} reports the reference results.
Appendices~\ref{app:environment}--\ref{app:verification} give the environment
profiles, task and chain definitions, verifier rules, stage records, and metric
calculations. Appendix~\ref{app:protocol} describes resets, source budgets,
target roles, seeds, artifact transfer, and paired reporting.
Appendix~\ref{app:reproduction} gives the memory formats, prompts, example
configuration, evaluation procedure, and reproduction commands. Run records
include the fixed task and verifier settings, seeds, source exposure, artifact
identities, and stage evidence needed to check the reported scores.

\bibliography{iclr2027_conference}
\bibliographystyle{iclr2027_conference}

\newpage
\appendix
\input{sections/Appendix.tex}

\end{document}

%% file: sections/math_commands.tex
\usepackage{amsmath,amsfonts,bm}

\def\eqref#1{equation~\ref{#1}}

\def\1{\bm{1}}

\DeclareMathAlphabet{\mathsfit}{\encodingdefault}{\sfdefault}{m}{sl}
\SetMathAlphabet{\mathsfit}{bold}{\encodingdefault}{\sfdefault}{bx}{n}



%% file: sections/Introduction.tex
\section{Introduction}

One of the central ambitions of autonomous intelligence is recursive self-improvement (RSI) \citep{chen2026recursive}: an agent should use experience from its own actions to become more capable over time. The goal is larger than learning one policy or storing one successful trajectory. An RSI system should turn experience into a capability, retain it, and apply it when the task, world, or required action changes. This loop is central to long-horizon autonomy: as tasks become harder, the agent must build on what it has already learned.

This view differs from evaluation settings in which training is separated from testing and improvement is summarized by a single score \citep{ren2026self}. For RSI, the important question is whether a gain remains useful after the original interaction has ended. A capability should survive a change in local conditions, help with a related goal, and combine with actions that were not part of the source task. RSI is therefore a systems problem: learning, memory, planning, and physical interaction must work together, while the resulting gain remains visible and attributable.

Embodied agents make this problem concrete and difficult. Their behavior depends on both internal state and physical state. A later success may reflect a tool left in the inventory, a changed world, a favorable location, or more source interaction rather than a reusable capability. A useful benchmark must therefore show that improvement survives a reset and appears in observable behavior.

\begin{center}
\begin{tikzpicture}
\node[draw=SkyBlue!85!black, fill=SkyBlue!8, rounded corners=3pt,
      line width=.45pt, inner xsep=5pt, inner ysep=5pt,
      text width=.94\linewidth, align=center, font=\footnotesize]
      {\textbf{\emph{When improved performance follows experience, does it reflect a reusable capability,\\or merely a changed physical state or execution condition?}}};
\end{tikzpicture}
\end{center}

We study this question through later behavior. A memory, skill file, or model update shows that an agent changed internally. It does not show that the agent gained a useful capability \citep{li2025optimus2,park2025mrsteve,li2026experience}. The evidence must link internal change to later action. It must also distinguish capability from favorable execution. To test that claim, we compare target behavior after a world reset and check both task outcomes and stage-level records.

Existing benchmarks address parts of this problem. MCU \citep{zheng2023mcu} evaluates diverse and combined Minecraft tasks with automatic verification. LifelongAgentBench \citep{zheng2025lifelongagentbench}, EvoAgentBench \citep{gao2026evoagentbench}, and PAST-Bench \citep{xue2026past} study learning across tasks, transfer, and persistent state in other settings. Embodied evaluation also needs to separate world resources from agent state, match source budgets, and record where a task chain fails \citep{ni2025embodied,cheng2026embodiedeval,luo2025robobench}. Otherwise, a score difference may reflect an easier world, more source interaction, or an early failure that hides later behavior.

Minecraft makes these controls practical. Collecting resources, crafting tools, and using them create visible task dependencies. A task-state graph represents these relations, and environment evidence verifies each stage. Resetting the world and inventory removes physical carryover, while stage records show progress before a chain stops \citep{guo2025mineworld,huang2025memory}.

We introduce MC-EvoBench to evaluate embodied RSI in Minecraft. Figure~\ref{fig:mechanism} shows its mechanism: measure progress, transfer only a declared agent artifact, and use matched controls to assess experience gains. Table~\ref{tab:rsi-comparison} compares this design with representative Minecraft benchmarks \citep{guss2019minerl,fan2022minedojo,hill2023mineplanner,wang2025openha}. MC-EvoBench provides a task catalog, verifiers, and three short probe families for wood-species crafting, wooden-tool transfer, and stone-tool transfer. The framework can study capability formation, transfer, retention, and composition. This paper focuses on forward transfer between graph-related tasks.

\begin{table*}[t]
\centering
\caption{Minecraft benchmark comparison for embodied RSI.}
\label{tab:rsi-comparison}
\scriptsize
\setlength{\tabcolsep}{3pt}
\renewcommand{\arraystretch}{1.15}
\resizebox{\textwidth}{!}{%
\begin{tabular}{lccccccc}
\toprule
\textbf{Dimension} & \shortstack{\textbf{MineRL}} & \textbf{MineDojo} & \textbf{MCU} & \textbf{MinePlanner} & \textbf{OpenHA} & \textbf{Voyager} & \cellcolor{tableblue}\textbf{MC-EvoBench} \\
\midrule
Capability evolution      & \rsiNo & \rsiNo & \rsiNo & \rsiNo & \rsiNo & \rsiYes & \cellcolor{tableblue}\rsiYes \\
Source isolation          & \rsiNo & \rsiNo & \rsiNo & \rsiNo & \rsiNo & \rsiNo & \cellcolor{tableblue}\rsiYes \\
Experience persistence    & \rsiNo & \rsiNo & \rsiNo & \rsiNo & \rsiNo & \rsiYes & \cellcolor{tableblue}\rsiYes \\
Capability transfer       & \rsiNo & \rsiNo & \rsiNo & \rsiNo & \rsiNo & \rsiYes & \cellcolor{tableblue}\rsiYes \\
Retention and composition & \rsiNo & \rsiNo & \rsiNo & \rsiNo & \rsiNo & \rsiNo & \cellcolor{tableblue}\rsiYes \\
Gain attribution          & \rsiNo & \rsiNo & \rsiNo & \rsiNo & \rsiNo & \rsiNo & \cellcolor{tableblue}\rsiYes \\
\bottomrule
\end{tabular}%
}
\vspace{-2pt}
\parbox{\textwidth}{\scriptsize \rsiYes\ Explicitly modeled or reported; \rsiNo\ not a primary design or evaluation dimension.}
\end{table*}

Our main contributions are as follows:

\begin{enumerate}
    \item \textbf{Embodied RSI as attributable capability change.} We formalize an agent as a policy, a persistent state, and an update rule, and separate this persistent state from the world state. A gain counts as experience-driven only when it appears after the target world is reset, only a declared artifact crosses the source--target boundary, and matched unrelated and shuffled sources rule out alternative explanations.
    \item \textbf{MC-EvoBench, an executable audit of experience-driven improvement.} It instantiates this definition in Minecraft with a task-state graph of 80 tasks, 111 states, and 369 typed edges organized into 19 chains; stage-level verifiers with three complementary metrics, Task Success Rate (TSR), Weighted Graph Progress (WGP), and Verified Milestone Progress (VMP); and a four-condition paired protocol with matched environment-step budgets and provenance records.
    \item \textbf{Controlled evidence on embodied transfer.} Controller-based agents clearly outperform planner-backed variants in direct capability. Under matched controls, the reference method shows no stable transfer gain, even though its related sources earn more verified progress and, on the longer probes, more computation, and early wood-collection and crafting bottlenecks leave later capabilities unobserved. Future RSI methods must therefore first clear these execution bottlenecks, and their gains should be claimed only after a controlled audit.
\end{enumerate}

Section~\ref{sec:related-work} reviews related benchmarks and agents. Section~\ref{sec:framework} defines the framework and benchmark. Section~\ref{sec:results} reports the reference results. Appendices~\ref{app:environment}--\ref{app:reproduction} give the task, environment, protocol, and reproduction details.

%% file: sections/Related_Work.tex
\section{Related Work}
\label{sec:related-work}

\paragraph{Minecraft and embodied-agent benchmarks.}
Minecraft combines visual control, persistent resources, and tasks with many prerequisites. MineDojo provides a large open-world task collection and a knowledge base that supports broad embodied-agent research \citep{fan2022minedojo}. BEDD extends MineRL \citep{guss2019minerl} with demonstrations and evaluation for tasks whose success can be difficult to describe with a single exact goal. MinePlanner focuses on long-horizon planning in large Minecraft worlds, while OpenHA studies open-source hierarchical agents and their execution in the same environment \citep{hill2023mineplanner, wang2025openha}. Voyager uses a curriculum, a skill library, and environment feedback for continued exploration and skill reuse, including in a new world \citep{wang2023voyager}. MCU evaluates a broad set of tasks, their combinations, and automatic assessment \citep{zheng2023mcu}. Together, these benchmarks cover open-world interaction, planning, task diversity, and automatic verification. They do not, however, isolate whether a controlled source interaction causes a later improvement after the target world is reset.

\paragraph{Lifelong learning and self-evolving agents.}
Several recent benchmarks study improvement across tasks and the use of stored experience. LifelongAgentBench tests skill gain, retention, and transfer across databases, operating systems, and knowledge graphs \citep{zheng2025lifelongagentbench}. EvoAgentBench uses agent traces to identify abilities and groups tasks by shared procedures, making transfer between related tasks explicit \citep{gao2026evoagentbench}. PAST-Bench compares runs with and without persistent state and tracks how experience is saved and reused in personal-agent settings \citep{xue2026past}. These studies establish useful ideas for measuring persistence and transfer, but their tasks do not expose the same physical-state confounds as embodied environments. In Minecraft, an apparent gain can instead come from a carried tool, a modified world, or a different starting location. MC-EvoBench addresses this boundary by resetting inventory and world changes while allowing only a declared agent artifact to transfer. Its task-state graph makes prerequisites and progress visible.

\paragraph{Memory-mediated improvement and benchmark position.}
Memory systems provide a practical mechanism for experience-driven improvement. Reflexion \citep{shinn2023reflexion} stores language feedback from failed attempts and uses it to guide later decisions. Agent Workflow Memory extracts reusable procedures from interaction traces for later tasks \citep{wang2026agent}. Skill libraries and model updates offer related ways to preserve information beyond a single episode. These methods show how an agent can store and retrieve experience, but reuse alone does not establish that a capability transferred. MC-EvoBench treats experience-driven improvement as an attribution problem. It separates source and target episodes, resets the physical state, and transfers only a declared artifact. Matched source conditions, verifier predicates, and stage records distinguish capability transfer from carried-over resources and partial execution.

%% file: sections/MC-EvoBench_Framework_and_Benchmark_Design.tex
\section{MC-EvoBench Framework and Benchmark Design}
\label{sec:framework}
\suppressfloats[t]

\begin{figure}[t]
\centering
\includegraphics[width=\linewidth]{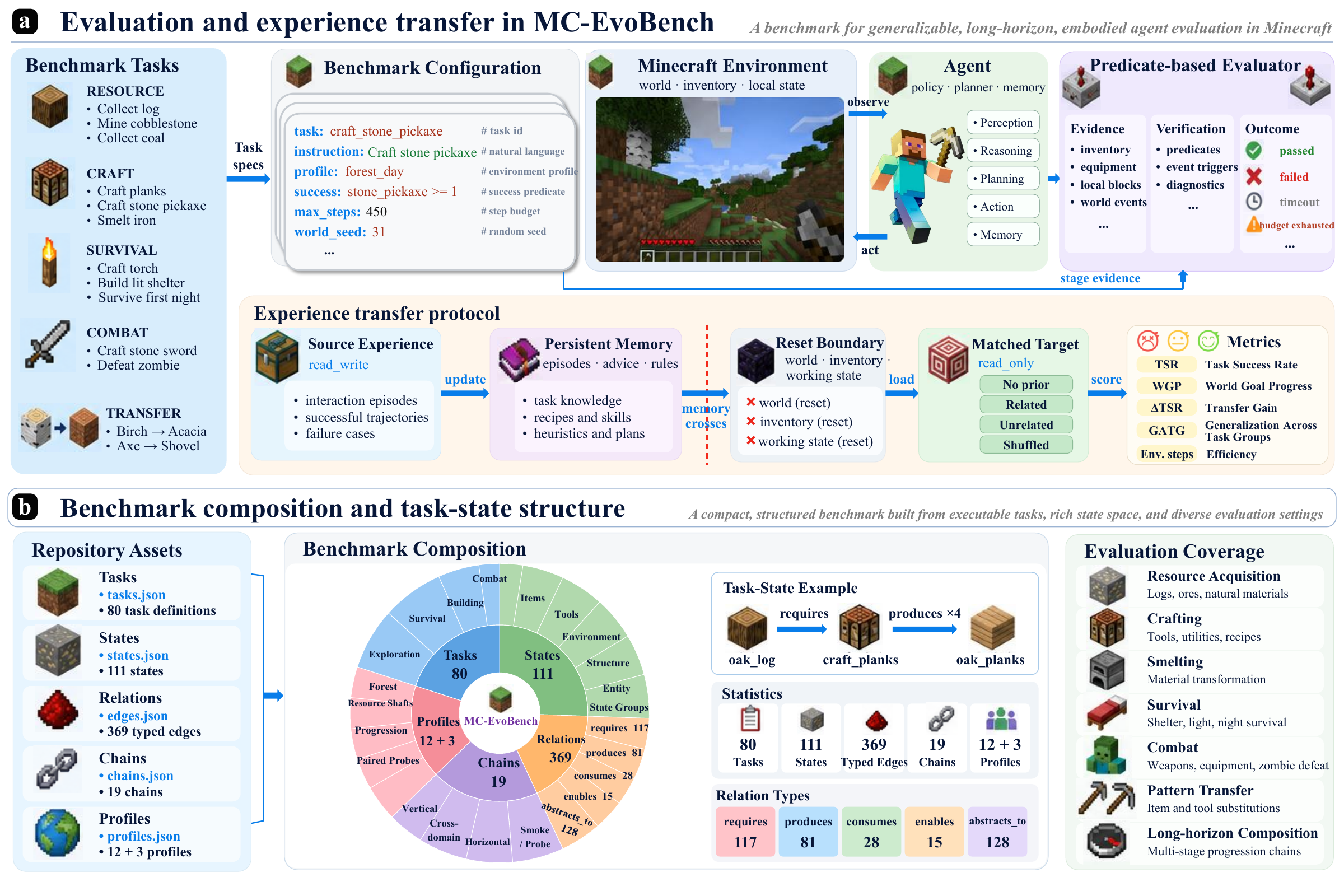}
\caption{Benchmark design of MC-EvoBench. Panel (a) follows a paired evaluation from source interaction and artifact formation to target-world reset, target execution, stage verification, and metric reporting. Panel (b) shows how the task catalog is organized into tasks, states, relations, chains, and profiles. These components support graph-based progress measurement, controlled transfer probes, and coverage of embodied capabilities across long-horizon task chains.}
\label{fig:benchmark-overview}
\end{figure}

MC-EvoBench tests whether experience changes later behavior. To support that claim, it resets the target world, checks outcomes against environment evidence, and compares matched source conditions. Figure~\ref{fig:benchmark-overview} shows how these parts work together. Panel (a) follows the agent from source interaction to target evaluation. Panel (b) shows the tasks, states, relations, chains, and environment profiles that define each test.

\subsection{Evaluation Target and State Boundary}
\label{sec:evaluation-target}

We ask whether controlled source interaction improves performance on a later target task in a reset world. A stored memory or model update cannot answer this question by itself. We represent an agent as \(A=(\pi,z,U)\): \(\pi\) selects actions from observations and goals, \(z\) is the state the method can retain, and \(U\) updates that state from experience. Source experience \(E_S\) produces \(z_S=U(z_0,E_S)\). The evaluator records and freezes \(z_S\) before the target run. No source resources, positions, world edits, route information, or success markers enter the target.

World state includes inventory, tools, resources, position, terrain, and world changes. Persistent agent state may include memories, prompts, replay data, checkpoints, or model updates. Every target run starts with the same registered world and inventory settings. Thus, source tools and world changes cannot explain a target score difference. A stored artifact provides evidence of a gain only when the target verifier also records a change in behavior.

\subsection{Task-State Graph}
\label{sec:task-state-graph}

MC-EvoBench represents task dependencies and observable resource flow with
\[
G=(\mathcal{T},\mathcal{S},\mathcal{E}),
\]
Here, \(\mathcal{T}\) contains tasks, \(\mathcal{S}\) contains observable states, and \(\mathcal{E}\) contains typed relations. States include items, tools, locations, structures, and task completion. The graph describes what tasks require, produce, or consume without prescribing an action sequence. These three relation types form the strict execution graph. The remaining types, \texttt{enables} and \texttt{abstracts\_to}, describe possible capability links and task categories; neither proves transfer. Appendix~\ref{app:tasks-2} defines each edge type and gives their counts.

Tasks form stage chains \(C=(\tau_1,\ldots,\tau_K)\). Each stage has a goal that the verifier checks against run evidence. The chain shows prerequisites and where execution stops; the agent remains free to choose its actions. Stage records reveal partial progress even when the final goal is not reached. Catalog and probe chains use the same task definitions and verifier rules. Reset settings, retained state, and step budgets are fixed before evaluation.

\subsection{Paired Protocol and Metrics}
\label{sec:protocol}
\label{sec:metrics}

Each comparison has a source phase and a target phase. The source agent interacts with registered tasks and forms \(z_S\). The evaluator then resets the target world and inventory, freezes \(z_S\), and runs the target task with the same verifier. Four target conditions serve different purposes. No-prior provides a baseline; related experience tests for a change; unrelated experience tests whether that change depends on task relevance; and shuffled related experience tests whether task order matters. We record source steps, termination, artifact identity, and target outcomes. Source episodes restart after early termination until the declared budget is used.

The verifier records the evidence for each stage, changes in resources, pass status, and why the run stopped. Let \(v_i\) indicate whether stage \(i\) passes. For each strict execution edge \(e\), \(w_e\) is its weight and \(q_e\) indicates verified progress. Let \(m_j\) indicate whether eligible milestone \(j\) is reached. The three direct metrics are
\begin{align}
\mathrm{TSR}&=\frac{\sum_{i=1}^{K}v_i}{K}, \label{eq:tsr}\\
\mathrm{WGP}&=\frac{\sum_{e\in E_{\mathrm{exec}}}w_e q_e}{\sum_{e\in E_{\mathrm{exec}}}w_e}, \label{eq:wgp}\\
\mathrm{VMP}&=\frac{\sum_{j=1}^{J}m_j}{J}. \label{eq:vmp}
\end{align}
TSR is the fraction of planned stages that pass; unreached stages remain in its denominator. WGP is the weighted fraction of strict graph edges supported by verified progress. VMP tracks eligible item or tool milestones observed before a chain ends. It captures partial progress, not full completion. We report all three metrics for each target condition and compare paired runs. Appendices~\ref{app:verification} and~\ref{app:protocol-6} give the verifier and reporting rules.

\subsection{Benchmark Instantiation}
\label{sec:benchmark}

MC-EvoBench implements this framework in Minecraft with a fixed task catalog, verifiers, probe profiles, and observation interface. A change to any of these components defines a new benchmark version. The same evaluation can therefore test different ways of storing experience, such as textual memories, skill libraries, prompts, or model updates, under a shared task and evidence specification.

\begin{table}[t]
\centering
\caption{Direct capability matrix over the 12 primary evaluation chains, grouped by method family. All entries are measured and reported as TSR / WGP / VMP in percent; the final group is the weighted average over all 12 chains.}
\label{tab:direct-capability}
\scriptsize
\setlength{\tabcolsep}{2.5pt}
\renewcommand{\arraystretch}{0.94}
\setlength{\aboverulesep}{1.5pt}
\setlength{\belowrulesep}{1.5pt}
\resizebox{\linewidth}{!}{%
\begin{tabular}{l*{12}{c}}
\toprule
\textbf{Method} & \multicolumn{3}{c}{\textbf{Core progression (5)}} & \multicolumn{3}{c}{\textbf{Catalog transfer (4)}} & \multicolumn{3}{c}{\textbf{Diversity (3)}} & \multicolumn{3}{c}{\textbf{Average (12)}}\\
\cmidrule(lr){2-4}\cmidrule(lr){5-7}\cmidrule(lr){8-10}\cmidrule(lr){11-13}
 & \textbf{TSR} & \textbf{WGP} & \textbf{VMP} & \textbf{TSR} & \textbf{WGP} & \textbf{VMP} & \textbf{TSR} & \textbf{WGP} & \textbf{VMP} & \textbf{TSR} & \textbf{WGP} & \textbf{VMP}\\
\midrule
\rowcolor{black!10}\multicolumn{13}{c}{\textit{Environment baseline}}\\
no-op & 3.5 & 1.5 & 0.0 & 0.0 & 0.0 & 0.0 & 0.0 & 0.0 & 0.0 & 1.5 & 0.6 & 0.0\\
\midrule
\rowcolor{black!10}\multicolumn{13}{c}{\textit{STEVE and JARVIS controller variants}}\\
STEVE-1 & 12.9 & 8.8 & 36.1 & 7.5 & 6.9 & 27.1 & 1.9 & 1.9 & 5.6 & 8.4 & 6.4 & 25.5\\
JARVIS-1 \textcolor{gray}{\fontsize{6}{7}\selectfont (controller-only)} & 9.9 & 5.7 & 26.7 & 8.7 & 8.0 & 37.5 & 1.9 & 1.9 & 5.6 & 7.5 & 5.5 & 25.0\\
\plannerline{JARVIS-1}{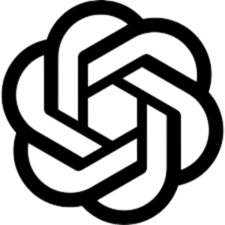} & 3.5 & 1.5 & 0.0 & 0.0 & 0.0 & 0.0 & 0.0 & 0.0 & 0.0 & 1.5 & 0.6 & 0.0\\
\plannerline{JARVIS-1}{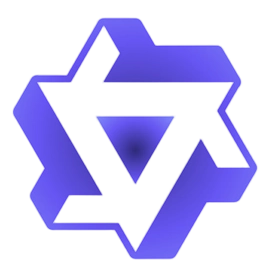} & 4.9 & 2.7 & 4.4 & 1.1 & 1.1 & 5.5 & 0.0 & 0.0 & 0.0 & 2.4 & 1.5 & 3.7\\
\plannerline{JARVIS-1}{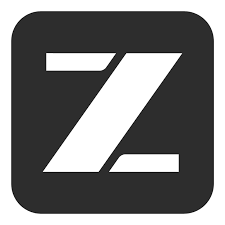} & 3.5 & 1.5 & 0.0 & 1.5 & 1.9 & 5.5 & 0.0 & 0.0 & 0.0 & 2.0 & 1.3 & 1.8\\
\plannerline{JARVIS-1}{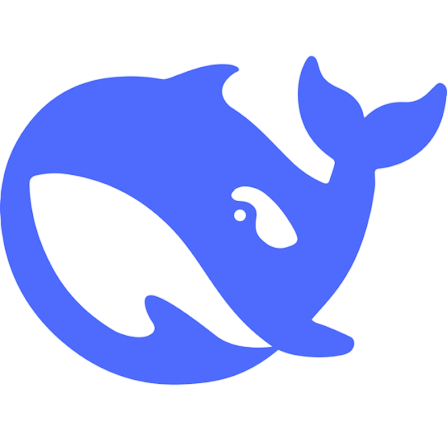} & 4.6 & 2.0 & 6.7 & 2.5 & 1.6 & 9.7 & 0.0 & 0.0 & 0.0 & 2.8 & 1.4 & 6.0\\
\plannerline{JARVIS-1}{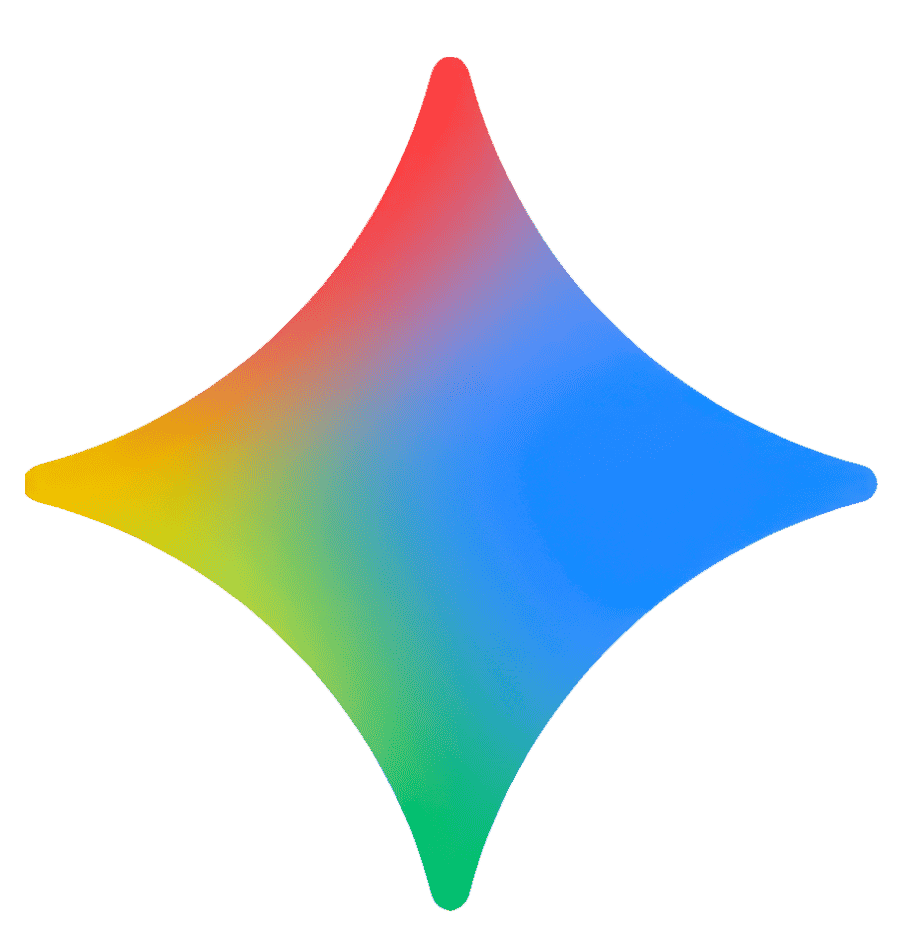} & 3.5 & 1.5 & 0.0 & 1.1 & 1.1 & 5.5 & 0.0 & 0.0 & 0.0 & 1.8 & 1.0 & 1.8\\
\midrule
\rowcolor{black!10}\multicolumn{13}{c}{\textit{Optimus variants}}\\
Optimus-1 \textcolor{gray}{\fontsize{6}{7}\selectfont (controller-only)} & 12.5 & 8.5 & 33.3 & 7.2 & 6.1 & 30.5 & 1.9 & 1.9 & 5.6 & 8.1 & 6.1 & 25.4\\
\plannerline{Optimus-1}{fig/GPT.png} & 6.7 & 4.1 & 11.7 & 1.7 & 1.3 & 5.5 & 0.0 & 0.0 & 0.0 & 3.4 & 2.1 & 6.7\\
\plannerline{Optimus-1}{fig/Qwen.png} & 6.7 & 4.1 & 11.7 & 3.2 & 3.2 & 11.1 & 0.0 & 0.0 & 0.0 & 3.9 & 2.8 & 8.6\\
\plannerline{Optimus-1}{fig/GLM.png} & 7.4 & 4.4 & 15.6 & 3.3 & 1.9 & 13.9 & 1.9 & 1.9 & 5.6 & 4.7 & 2.9 & 12.5\\
\plannerline{Optimus-1}{fig/deepseek.png} & 9.2 & 5.7 & 19.5 & 3.2 & 3.2 & 11.1 & 0.0 & 0.0 & 0.0 & 4.9 & 3.4 & 11.8\\
\plannerline{Optimus-1}{fig/Gemini.png} & 6.7 & 4.1 & 11.7 & 3.3 & 1.9 & 13.9 & 0.0 & 0.0 & 0.0 & 3.9 & 2.3 & 9.5\\
Optimus-3 action path & 3.5 & 1.5 & 0.0 & 0.0 & 0.0 & 0.0 & 0.0 & 0.0 & 0.0 & 1.5 & 0.6 & 0.0\\
\midrule
\rowcolor{black!10}\multicolumn{13}{c}{\textit{VPT variants}}\\
VPT \textcolor{gray}{\fontsize{6}{7}\selectfont (STEVE-1 checkpoint)} & 4.2 & 1.8 & 3.3 & 1.5 & 1.9 & 5.5 & 0.0 & 0.0 & 0.0 & 2.3 & 1.4 & 3.2\\
VPT \textcolor{gray}{\fontsize{6}{7}\selectfont (official 2x foundation weights)} & 4.9 & 2.1 & 6.7 & 2.3 & 0.9 & 12.5 & 5.2 & 5.2 & 13.9 & 4.1 & 2.5 & 10.4\\
\bottomrule
\end{tabular}%
}
\end{table}

\subsubsection{Frozen Catalog and Probe Families}
\label{sec:catalog}

The catalog contains 80 tasks, 111 state nodes, 369 graph edges, 19 registered chains, 12 general profiles, and 3 paired-transfer profiles. Its 19 chains include 12 primary evaluation chains and 7 calibration or probe chains. Tasks span building, survival, exploration, and combat. Long chains test execution across many prerequisites. Short probes test local transfer that an early failure in a long chain could hide. The probes use catalog tasks and the same verifier rules. Appendices~\ref{app:suite} and~\ref{app:environment-3} list the tasks, chains, and profiles.

The three probe families and their budgets are listed in Table~\ref{tab:app-d2} of Appendix~\ref{app:protocol-3}. In each family, related and target chains share a dependency pattern but end with a different object or tool. This lets us test transfer while keeping the required preparation steps. Each stage has a 450-step cap. After early termination, a new episode starts until the family budget is used.

\subsubsection{Environment and Observation Interface}
\label{sec:environment}

Paired-transfer runs use public world seed 31, an empty target inventory, and one of three fixed environment profiles (Appendix~\ref{app:environment-3}); direct-capability runs use the seeds and per-stage profiles in Table~\ref{tab:app-direct-protocol}. Profiles set time, weather, spawn, clearance, and public resources. They do not provide target items, routes, private coordinates, or verifier state. The agent must observe and collect resources through interaction. Each target condition starts with a world and inventory reset; only the declared agent artifact can transfer. The agent sees normal observations, while the graph, verifier rules, diagnostics, and hidden coordinates remain unavailable. Appendices~\ref{app:environment-2} and~\ref{app:verification-1} detail these access rules and verifier checks.

%% file: sections/Experimental_Results.tex
\section{Experimental Results}
\label{sec:results}

We report two complementary views of the benchmark. The direct capability matrix measures immediate performance across the registered catalog without a source--target transfer claim. The paired probes then test whether a declared source artifact changes behavior on a later target after the world is reset. Keeping these views separate distinguishes basic execution ability from experience-driven improvement.

\begin{table}[t]
\centering
\caption{Paired transfer results for the three probe families. Values are target-condition means over five matched seeds; each cell lists no-prior (N), related (R), unrelated (U), and shuffled (S) conditions in that order. TSR, WGP, and VMP measure stage completion, graph progress, and verified milestone progress, respectively, after the target world is reset.}
\label{tab:main-results}
\scriptsize
\setlength{\tabcolsep}{2pt}
\renewcommand{\arraystretch}{1.12}
\begin{tabular}{@{}p{0.6in}>{\centering\arraybackslash}p{1.39in}>{\centering\arraybackslash}p{1.39in}>{\centering\arraybackslash}p{1.39in}@{}}
\toprule
\rowcolor{tableblue}
\textbf{Probe family} & \multicolumn{3}{c}{\textbf{Target-condition means}} \\
\cmidrule(lr){2-4}
 & \textbf{TSR} {\textcolor{gray}{(N/R/U/S)}} & \textbf{WGP} {\textcolor{gray}{(N/R/U/S)}} & \textbf{VMP} {\textcolor{gray}{(N/R/U/S)}} \\
\midrule
\rowcolor{black!3}
Wood species & 0.400 / 0.400 / 0.200 / 0.500 & 0.360 / 0.400 / 0.200 / 0.480 & 0.400 / 0.400 / 0.200 / 0.500 \\
Wooden tool & 0.320 / 0.200 / 0.160 / 0.240 & 0.225 / 0.150 / 0.100 / 0.167 & 0.560 / 0.367 / 0.333 / 0.467 \\
\rowcolor{black!3}
Stone tool & 0.150 / 0.050 / 0.025 / 0.075 & 0.082 / 0.021 / 0.010 / 0.031 & 0.533 / 0.200 / 0.100 / 0.300 \\
\bottomrule
\end{tabular}
\end{table}

\subsection{Evaluation Setup}
\label{sec:setup}

The paired evaluation uses three probe families: wood-species crafting, wooden-tool transfer, and stone-tool transfer. Each family uses five paired agent seeds and four target conditions: no prior, related experience, unrelated experience, and shuffled related experience. Every target starts from the same world and inventory state for its family. Only the declared agent artifact may cross the source--target boundary. We report TSR, WGP, and VMP as defined in Section~\ref{sec:metrics}. Source exposure is matched within each family; planner calls and token usage are recorded separately because equal environment exposure does not imply equal computation.

\subsection{Direct Capability Matrix}
\label{sec:direct-capability}

The direct matrix covers 12 primary evaluation chains (five core-progression, four catalog-transfer, and three diversity) and nine direct configurations, each evaluated with three seeds. Each cell in Table~\ref{tab:direct-capability} averages TSR / WGP / VMP over the chains in a partition; the final columns weight the partitions by their sizes (5/4/3), and Appendix Table~\ref{tab:app-direct-detail} gives per-chain scores. These runs measure direct capability rather than transfer; ExpeL \citep{zhao2024expel} has one separately marked report per chain and is excluded from these aggregates.

Controller-based execution clearly outperforms planner-backed execution. STEVE-1 \citep{lifshitz2023steve} and the controller-only JARVIS-1 \citep{wang2024jarvis} and Optimus-1 \citep{li2024optimus} variants reach average TSR/WGP/VMP of 8.4/6.4/25.5, 7.5/5.5/25.0, and 8.1/6.1/25.4, with the largest margins on the core and catalog-transfer partitions, where progress requires several dependent actions. The no-op baseline and the GPT-planned JARVIS-1 remain at the execution floor, confirming that the benchmark separates active task solving from background behavior. No planner dominates all three metrics: DeepSeek gives the highest average TSR for both JARVIS-1 (2.8) and Optimus-1 (4.9), whereas the highest VMP comes from DeepSeek for JARVIS-1 (6.0) but from GLM for Optimus-1 (12.5). The official VPT \citep{baker2022video} foundation weights improve over the STEVE-1 checkpoint (average VMP 10.4 versus 3.2) but remain below the controller variants.

\finding{Capability is first bounded by execution.}{Controllers consistently outperform planner-backed variants, and no planner leads on all three metrics.}

\subsection{Paired Transfer Results}
\label{sec:main-results}

Table~\ref{tab:main-results} reports mean target scores in the order no-prior / related / unrelated / shuffled. The study contains 105 source and target role reports in total, with 60 target reports contributing to the four-condition comparison.

For wood-species crafting, related experience gives scores close to no prior. The shuffled control scores at least as high on TSR and WGP. For wooden-tool and stone-tool transfer, related experience scores below no prior on all three metrics. On the stone-tool probe, related TSR, WGP, and VMP are 0.050, 0.021, and 0.200, compared with 0.150, 0.082, and 0.533 for no prior.

\finding{Stored experience is not acquired capability.}{After a world reset with artifact-only transfer, related experience does not reliably exceed no prior and lowers all three metrics on both tool probes.}

\subsection{Paired Contrasts and Uncertainty}
\label{sec:paired-contrasts}

For each direct metric, let $M_0$, $M_R$, $M_U$, and $M_S$ denote the no-prior, related, unrelated, and shuffled target scores. We report the related-source change $\Delta R=M_R-M_0$. For completeness, Table~\ref{tab:transfer-contrasts} also reports the adjusted contrast
\[
\mathrm{GATG}=M_R-\frac{1}{2}(M_U+M_S).
\]
GATG is used here only as a descriptive paired contrast; TSR, WGP, and VMP remain the benchmark's direct metrics. Intervals are two-sided 95\% Student-$t$ intervals over the five paired seeds.

\begin{table}[t]
\centering
\caption{Paired transfer contrasts over five matched seeds. Columns are grouped by probe family and report the related-source change ($\Delta R$), the diagnostic GATG contrast, and its two-sided 95\% Student-$t$ confidence interval. GATG is not an additional benchmark metric.}
\label{tab:transfer-contrasts}
\scriptsize
\setlength{\tabcolsep}{2.5pt}
\renewcommand{\arraystretch}{1.10}
\resizebox{\linewidth}{!}{%
\begin{tabular}{@{}l*{9}{c}@{}}
\toprule
\rowcolor{tableblue}
\textbf{Metric} & \multicolumn{3}{c}{\textbf{Wood species}} & \multicolumn{3}{c}{\textbf{Wooden tool}} & \multicolumn{3}{c}{\textbf{Stone tool}}\\
\cmidrule(lr){2-4}\cmidrule(lr){5-7}\cmidrule(lr){8-10}
 & $\Delta R$ & \textbf{GATG} & \textbf{95\% CI} & $\Delta R$ & \textbf{GATG} & \textbf{95\% CI} & $\Delta R$ & \textbf{GATG} & \textbf{95\% CI}\\
\midrule
\rowcolor{black!3}
TSR & 0.000 & 0.050 & [-0.723, 0.823] & -0.120 & 0.000 & [-0.152, 0.152] & -0.100 & 0.000 & [-0.095, 0.095]\\
WGP & 0.040 & 0.060 & [-0.718, 0.838] & -0.075 & 0.017 & [-0.132, 0.166] & -0.062 & 0.000 & [-0.039, 0.039]\\
VMP & 0.000 & 0.050 & [-0.723, 0.823] & -0.193 & -0.033 & [-0.248, 0.181] & -0.333 & 0.000 & [-0.380, 0.380]\\
\bottomrule
\end{tabular}
}%
\end{table}

Table~\ref{tab:transfer-contrasts} separates the direct related-source change from the control-adjusted contrast. Relative to no prior, related experience improves only WGP on the wood-species probe ($\Delta R=+0.040$); TSR and VMP are unchanged. The two tool-transfer probes show negative changes on all three metrics, with the largest decrease in VMP for stone-tool transfer ($\Delta R=-0.333$). After adjusting for the unrelated and shuffled controls, the wood-species probe retains a small positive GATG on TSR, WGP, and VMP ($+0.050$, $+0.060$, and $+0.050$). The wooden-tool probe is mixed: GATG is near zero for TSR and slightly positive for WGP, but negative for VMP. All reported confidence intervals include zero, and the intervals are wide because they are estimated from five matched seeds.

\finding{Self-improvement cannot be claimed without controls.}{Once unrelated and shuffled sources are accounted for, the largest contrast is only $+0.06$ and every 95\% interval includes zero.}

\subsection{Source Cost and Run Records}
\label{sec:source-cost}

Matching environment steps does not equalize episodes or planner calls. Table~\ref{tab:source-cost} reports the observed source cost: step budgets are fixed within each family (900, 2,250, and 3,600), while episode counts, LLM calls, and termination states depend on how each run proceeds.

\begin{table}[t]
\centering
\caption{Observed source-side cost under matched family-level exposure. Values are mean $\pm$ standard deviation where applicable; environment steps are fixed within each probe family, while episode counts, source TSR, planner calls, and termination states are recorded from the runs.}
\label{tab:source-cost}
\newcommand{\msd}[2]{$#1$\,{\tiny\color{gray}$\pm#2$}}
\scriptsize
\setlength{\tabcolsep}{2.4pt}
\renewcommand{\arraystretch}{1.08}

\begin{tabular}{@{}llrrrrl@{}}
\toprule
\rowcolor{tableblue}
\textbf{Family} & \textbf{Source} & \multicolumn{2}{c}{\textbf{Execution budget}} & \multicolumn{2}{c}{\textbf{Observed cost}} & \textbf{Termination summary}\\
\cmidrule(lr){3-4}\cmidrule(lr){5-6}
 &  & \textbf{Env. steps} & \textbf{Episodes} & \textbf{Source TSR} & \textbf{LLM calls} & \\
\midrule
\rowcolor{black!3}
Wood species & Related & \msd{900}{0} & \msd{2.00}{0.00} & \msd{0.100}{0.224} & \msd{4.40}{0.89} & \texttt{budget\_exhausted}=1; \texttt{passed}=1; \texttt{timeout}=9\\
Wood species & Unrelated & \msd{900}{0} & \msd{2.60}{0.89} & \msd{0.000}{0.000} & \msd{5.80}{2.68} & \texttt{budget\_exhausted}=2; \texttt{passed}=6; \texttt{timeout}=8\\
Wood species & Shuffled & \msd{900}{0} & \msd{2.00}{0.00} & \msd{0.000}{0.000} & \msd{4.00}{0.00} & \texttt{timeout}=10\\
\midrule
Wooden tool & Related & \msd{2{,}250}{0} & \msd{5.00}{0.71} & \msd{0.080}{0.110} & \msd{14.80}{1.10} & \texttt{budget\_exhausted}=5; \texttt{passed}=12; \texttt{timeout}=20\\
Wooden tool & Unrelated & \msd{2{,}250}{0} & \msd{6.00}{0.00} & \msd{0.000}{0.000} & \msd{12.00}{0.00} & \texttt{budget\_exhausted}=5; \texttt{timeout}=25\\
Wooden tool & Shuffled & \msd{2{,}250}{0} & \msd{5.00}{0.00} & \msd{0.000}{0.000} & \msd{10.00}{0.00} & \texttt{timeout}=25\\
\midrule
\rowcolor{black!3}
Stone tool & Related & \msd{3{,}600}{0} & \msd{8.20}{0.84} & \msd{0.150}{0.105} & \msd{24.80}{4.60} & \texttt{budget\_exhausted}=5; \texttt{passed}=21; \texttt{timeout}=36\\
Stone tool & Unrelated & \msd{3{,}600}{0} & \msd{8.00}{0.00} & \msd{0.000}{0.000} & \msd{16.40}{0.89} & \texttt{budget\_exhausted}=1; \texttt{passed}=1; \texttt{timeout}=39\\
Stone tool & Shuffled & \msd{3{,}600}{0} & \msd{8.00}{0.00} & \msd{0.000}{0.000} & \msd{16.00}{0.00} & \texttt{timeout}=40\\
\bottomrule
\end{tabular}

\end{table}

The related condition is the only one with nonzero source TSR (0.100, 0.080, and 0.150) and accounts for 34 of the 41 \texttt{passed} records, whereas shuffled runs end only by timeout. It also uses more planner calls on the two longer probes, reaching $24.80\pm4.60$ calls for stone-tool transfer versus 16.40 and 16.00 for the unrelated and shuffled controls.

\begin{figure}[t]
\centering
\includegraphics[width=\linewidth]{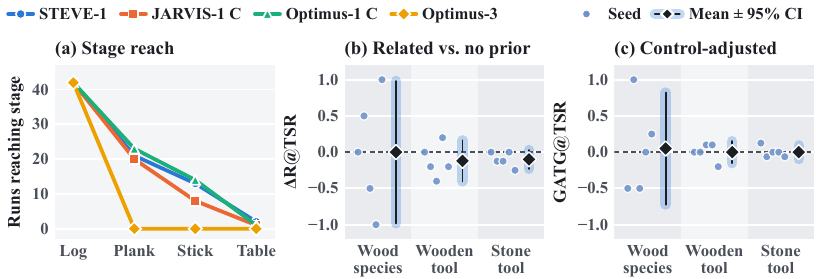}
\caption{Failure-stage and per-seed diagnostics. (a) Runs, out of 42 attempts per configuration, reaching each wood and crafting stage. (b) Related-source $\Delta R$ and (c) control-adjusted GATG on TSR; dots are matched seeds; diamonds with shaded bars give the mean and 95\% Student-$t$ interval.}
\label{fig:stage-seed}
\end{figure}

\finding{More experience alone does not yield transfer.}{On the longer probes, related sources gain more verified progress and planner calls than the controls, yet this extra experience yields no target gain.}

\subsection{Failure Stages, Seed Variation, and Scope}
\label{sec:failure-diagnostics}
\label{sec:interpretation}

In the failure-stage audit of the 57 direct-matrix reports per configuration (Figure~\ref{fig:stage-seed}a), attrition concentrates in the first crafting steps: of the 42 reports that include wood collection, the three controller variants bring 20--23 to plank crafting, at most 14 to sticks, and at most 2 to the crafting table, while the Optimus-3 \citep{li2025optimus} action path never passes the initial wood stage. A low final score can therefore reflect an early execution bottleneck rather than the failure of a later capability, which remains unobserved. Per-seed effects (\mbox{Figure~\ref{fig:stage-seed}b--c}) vary most for wood species, where the small positive mean GATG ($+0.050$) is driven by a single seed ($+1.0$) and falls to $-0.19$ without it; the tool probes vary less and have GATG means near zero. All three intervals cross zero, consistent with Table~\ref{tab:transfer-contrasts}.

Taken together, the direct matrix measures immediate capability rather than transfer, and the paired probes should be read against its low execution floor. Under the current public instances, seeds, budgets, and reference method, related experience does not produce a stable positive transfer effect; this does not rule out gains under a different agent, artifact, task distribution, or source budget. Appendices~\ref{app:verification-3}, \ref{app:protocol-6}, and~\ref{app:agents-4} give the stage records, paired-reporting rules, and configuration fields needed to reproduce these comparisons.

\finding{Unobserved capabilities cannot be evaluated.}{Most runs stop in the first crafting stages and the wood-species gain hinges on one seed, so progress needs stage-level evidence, not final success.}

%% file: sections/Appendix.tex
\setlength{\parindent}{0pt}\setlength{\parskip}{4pt}\setlength{\emergencystretch}{2em}
\widowpenalty=10000\clubpenalty=10000\setlength{\LTpre}{2pt}\setlength{\LTpost}{3pt}\fvset{numbersep=5pt}\renewcommand{\theFancyVerbLine}{\textcolor{black!50}{\fontsize{6}{7}\selectfont\arabic{FancyVerbLine}}}
\tcbset{
  colback=black!2,
  colframe=black!35,
  colbacktitle=black!8,
  coltitle=black,
  fonttitle=\bfseries\fontsize{9}{11}\selectfont,
  fontupper=\fontsize{9}{11}\selectfont,
  boxrule=0.45pt,
  arc=0pt,
  left=7pt,right=7pt,top=5pt,bottom=5pt,
  toptitle=5pt,bottomtitle=5pt,
  before skip=7pt,after skip=7pt
}

\Needspace{6\baselineskip}
\newcounter{apptable}[section]
\renewcommand{\theapptable}{\thesection\arabic{apptable}}
\newcounter{appfigure}[section]
\renewcommand{\theappfigure}{\thesection\arabic{appfigure}}
\numberwithin{equation}{section}
\newcounter{appcard}[section]
\renewcommand{\theappcard}{\thesection\arabic{appcard}}
\section{Minecraft Environment Settings}
\label{app:environment}
MC-EvoBench evaluates whether interaction experience can improve subsequent embodied execution. Its experimental unit joins a source experience, a persistent agent artifact, and a separately initialized target evaluation. This appendix specifies the environment, task construction, verification, experimental protocol, agent interfaces, task suite, and reproducibility materials.\par
\Needspace{2\baselineskip}
\subsection{Game world and interaction interface}
\label{app:environment-1}
The primary environment is Minecraft 1.16.5 through MineRL, with a custom survival specification named \texttt{MCBench\allowbreak{}Human\allowbreak{}Survival-\allowbreak{}v0}. It extends the HumanSurvival environment with observations for inventory, local resources, world conditions, and task events. The agent interacts with the game through an adapter, while a separate evaluator converts the environment response into a normalized observation snapshot. This separation supports a common task definition across agents with different planning and control architectures.\par
The native control interface consists of movement, camera rotation, attack, use, and inventory operations. Camera commands change pitch and yaw; the meaning of attack and use depends on whether the agent is interacting with the world or an inventory interface. These environment actions produce the observations used for task verification. Text-conditioned controllers translate a task instruction into this action interface. A programmatic agent may use a different bridge, in which case its action granularity is reported with the method settings in Appendix~\ref{app:agents}.\par
\Needspace{2\baselineskip}
\refstepcounter{apptable}\label{tab:app-a1}
{\small Table \theapptable. Interaction and observation interfaces. The evaluator consumes the evidence needed by a task; an agent consumes the modalities exposed by its declared adapter.\par}\nopagebreak\smallskip
{\fontsize{9}{10.7}\selectfont\setlength{\tabcolsep}{2pt}\renewcommand{\arraystretch}{1.10}
\noindent\begin{minipage}{\linewidth}\begin{tabular}{>{\raggedright\arraybackslash}p{88.26pt}>{\raggedright\arraybackslash}p{134.32pt}>{\raggedright\arraybackslash}p{161.18pt}}
\toprule
\textbf{Component} & \textbf{Representation} & \textbf{Role}\\\midrule
Visual input & First-person RGB frame & Scene perception and low-level control\\
Movement & Forward, back, left, right; jump, sneak, sprint & Navigation and locomotion\\
Camera & Relative pitch and yaw & Viewpoint and interaction targeting\\
Interaction & Attack, use, drop, inventory, hotbar selection & Resource collection, construction, and item manipulation\\
Inventory and equipment & Item counts; normalized equipped-item names & Agent context when exposed; task verification\\
Local resource evidence & Ray observation and block-count grid & Structured perception or evaluator evidence, according to the adapter\\
World and mission evidence & Time, light, survival state, entity-kill counters & Event verification and execution diagnostics\\
\bottomrule\end{tabular}\end{minipage}\par}
\smallskip
\par\smallskip\noindent\begin{minipage}{\linewidth}
\centering\includegraphics[width=0.76\linewidth]{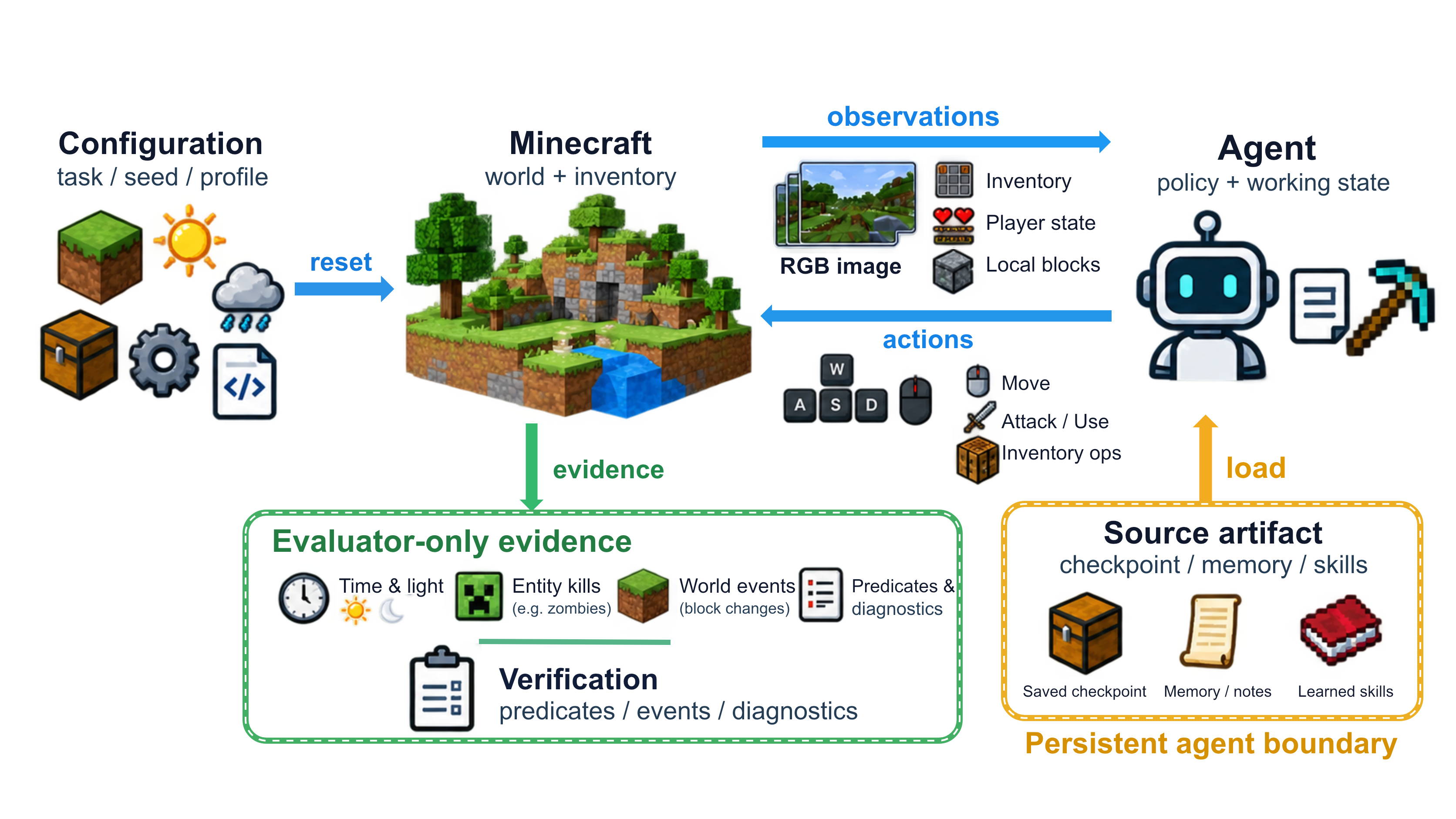}\par\smallskip
\refstepcounter{appfigure}\label{fig:app-a1}
{\small\raggedright Figure \theappfigure. The task specification connects initialization, agent execution, and evaluation. World configuration fixes the problem instance. Agent inputs follow the selected interface, while success is evaluated from environment evidence. Persistent experience crosses source and target runs through the artifact boundary defined in Appendix~\ref{app:protocol}.\par}
\end{minipage}\par\smallskip
\Needspace{2\baselineskip}
\subsection{Observation access and normalized evidence}
\label{app:environment-2}
The catalog distinguishes \texttt{native\_\allowbreak{}perception} and \texttt{structured\_\allowbreak{}state} observation sources. Each source records a path, its primary or diagnostic role, its interpretation, and an access declaration. Inventory, equipment, and basic survival measurements are examples of declared agent-visible sources. Local block counts, mission flags, kill counters, and world-light diagnostics are classified as structured evidence. RGB-based and structured-state methods therefore remain distinguishable even when they share the same task verifier.\par
At runtime, the snapshot groups observations into inventory, equipment, player statistics, counters, flags, raw observations, and environment metadata. Equipment names are normalized across MineRL representations, and raw entity-kill observations can populate the corresponding normalized counter. The evaluator preserves the raw channel alongside the normalized value, which makes it possible to trace a predicate back to its observation source. The actual model inputs are determined by the adapter; Appendix~\ref{app:agents} describes the built-in interfaces.\par
The local block grid spans offsets from $(-3,-2,-3)$ to $(3,2,3)$ around the player, giving a $7\times5\times7$ observation volume. The ray channel supplies a block identity and distance. Grid counts support local structure checks; ray observations support direct resource visibility. Profile-provided resource markers describe the configured scene and serve as diagnostics. The coal-discovery predicate uses coal observed in the ray channel or local block grid.\par
\Needspace{2\baselineskip}
\subsection{Environment profiles and controlled resources}
\label{app:environment-3}
An environment profile specifies a world seed, time of day, weather, initial inventory, and optional placement metadata. The benchmark includes 12 general environment profiles, all with empty initial inventories. Four forest profiles provide progression, daytime, nighttime, and dusk settings. Controlled-resource profiles place scarce materials near the spawn so that a finite evaluation budget can probe tool preparation, mining, collection, and downstream use. Ore is a world object that the agent must interact with.\par
\Needspace{2\baselineskip}
\refstepcounter{apptable}\label{tab:app-a2}
{\small Table \theapptable. General environment profile registry. Codes are used in Appendix~\ref{app:suite}. Seeds are profile defaults; an experiment can override them explicitly. All entries initialize an empty inventory.\par}\nopagebreak\smallskip
{\fontsize{9}{10.7}\selectfont\setlength{\tabcolsep}{2pt}\renewcommand{\arraystretch}{1.10}
\noindent\begin{minipage}{\linewidth}\begin{tabular}{>{\raggedright\arraybackslash}p{30.31pt}>{\raggedright\arraybackslash}p{151.54pt}>{\raggedright\arraybackslash}p{30.31pt}>{\raggedright\arraybackslash}p{166.69pt}}
\toprule
\textbf{Code} & \textbf{Profile} & \textbf{Seed} & \textbf{Time / controlled setting}\\\midrule
P01 & \texttt{forest\_\allowbreak{}progression} & 42 & Tick 1,000; time progresses\\
P02 & \texttt{forest\_\allowbreak{}day} & 7 & Tick 1,000; daylight\\
P03 & \texttt{forest\_\allowbreak{}night} & 11 & Tick 14,000; hostile spawning enabled\\
P04 & \texttt{forest\_\allowbreak{}dusk} & 9 & Tick 12,000; dusk\\
P05 & \texttt{fixed\_\allowbreak{}coal\_\allowbreak{}shaft} & 21 & Exposed coal; radius 4; catalog y-band 73--74\\
P06 & \texttt{fixed\_\allowbreak{}iron\_\allowbreak{}shaft} & 21 & Iron; radius 4; catalog y-band 70--71\\
P07 & \texttt{fixed\_\allowbreak{}gold\_\allowbreak{}shaft} & 21 & Gold; radius 4; catalog y-band 70--71\\
P08 & \texttt{fixed\_\allowbreak{}redstone\_\allowbreak{}shaft} & 21 & Redstone; radius 4; catalog y-band 70--71\\
P09 & \texttt{fixed\_\allowbreak{}lapis\_\allowbreak{}shaft} & 21 & Lapis; radius 4; catalog y-band 70--71\\
P10 & \texttt{fixed\_\allowbreak{}diamond\_\allowbreak{}shaft} & 21 & Diamond; radius 4; catalog y-band 70--71\\
P11 & \texttt{fixed\_\allowbreak{}emerald\_\allowbreak{}shaft} & 21 & Emerald; radius 4; catalog y-band 70--71\\
P12 & \texttt{fixed\_\allowbreak{}ore\_\allowbreak{}progression\_\allowbreak{}shaft} & 21 & Seven-resource gallery; radius 6; catalog y-band 70--71\\
\bottomrule\end{tabular}\end{minipage}\par}
\smallskip
Resource placement is specified relative to the configured or actual spawn, as recorded in the profile. The multi-resource gallery contains coal, iron, gold, redstone, lapis, diamond, and emerald. These local arrangements standardize access to task materials; they measure execution under controlled resource availability. The source files retain block placement, facing direction, resource identity, and allowed-task metadata.\par
The paired-transfer configurations use three additional profiles: a mixed wood-species grove, an oak grove, and a stone gallery. They use world seed 31, daylight, clear weather, disabled spawning, and an empty initial inventory. Their nearby resources support paired source and target tasks with comparable access conditions. Related and unrelated source content, interaction budgets, and target substitutions are specified together in Appendix~\ref{app:protocol}.\par
\Needspace{2\baselineskip}
\subsection{State continuity within a task chain}
\label{app:environment-4}
The chain specification selects either a task-level or chain-level profile policy. Under the task-level policy, the runner selects each task's profile; with the default per-stage reset setting, a profile change causes an environment reset. Under the chain-level policy, one selected profile governs the chain. Consecutive stages that remain in the same environment can reuse world resources and inventory, making state continuity an explicit part of chain evaluation.\par
This within-chain continuity is separate from the source-to-target boundary. A target transfer evaluation starts from its prescribed environment state and receives only the permitted persistent artifact. A tool retained between two stages changes their execution conditions; a tool reconstructed after a target reset provides evidence about execution under the retained agent state. The protocol records these two situations separately.\par

\clearpage
\section{Task Construction and Composition}
\label{app:tasks}
\Needspace{5\baselineskip}
\subsection{Task specification and benchmark coverage}
\label{app:tasks-1}
A task is a compact executable contract: an identifier, an instruction, a domain, a step budget, an environment profile, a success specification, and observation-source declarations. Instructions describe ordinary Minecraft objectives such as collecting a resource or crafting a tool. The success specification supplies the measurable endpoint. Keeping these fields together allows the same task to be placed in a short probe, a longer chain, or an independent evaluation without redefining its meaning.\par
The task catalog contains 80 tasks across building, survival, exploration, and combat. The domain counts are 40, 26, 7, and 7, respectively. The accompanying graph contains 111 state nodes and 369 edges. Its composition rules connect resource acquisition, crafting, tool use, environmental conditions, and high-level events. Appendix~\ref{app:suite} provides the complete task and chain definitions; the machine-readable tables retain the original instructions and predicates.\par
\Needspace{5\baselineskip}
\refstepcounter{apptable}\label{tab:app-b1}
{\small Table \theapptable. Catalog partitions and their execution denominators. A task identifier can occur more than once in a chain or across chains. Stage occurrences are therefore counted separately from distinct task membership.\par}\nopagebreak\smallskip
{\fontsize{9}{10.7}\selectfont\setlength{\tabcolsep}{2pt}\renewcommand{\arraystretch}{1.10}
\noindent\begin{minipage}{\linewidth}\begin{tabular}{>{\raggedright\arraybackslash}p{82.26pt}>{\raggedright\arraybackslash}p{44.87pt}>{\raggedright\arraybackslash}p{33.65pt}>{\raggedright\arraybackslash}p{52.35pt}>{\raggedright\arraybackslash}p{160.79pt}}
\toprule
\textbf{Partition} & \textbf{Declared tasks} & \textbf{Chains} & \textbf{Stage occurrences} & \textbf{Purpose}\\\midrule
Core progression & 33 & 5 & 71 & Resource, tool-tier, survival, and combat progression\\
Transfer & 31 & 4 & 50 & Item, recipe, and schema substitutions\\
Supplementary diversity & 16 & 3 & 19 & Breadth of collection and resource interaction\\
Calibration and probe & 0 & 7 & 51 & Execution and verifier checks\\
Primary evaluation total & 80 & 12 & 140 & Core, transfer, and diversity combined\\
\bottomrule\end{tabular}\end{minipage}\par}
\smallskip
The first three partitions define the primary evaluation suite. The calibration partition reuses task definitions for engineering checks and is reported separately. This organization supplies both a stable comparison set and a small set of diagnostic entry points. The task-membership split describes the catalog taxonomy; shared prerequisites can still recur in the executable chains.\par
\Needspace{5\baselineskip}
\subsection{Task--state graph}
\label{app:tasks-2}
\Needspace{7\baselineskip}
Let $\mathcal{T}$ denote task nodes and $\mathcal{S}$ state nodes. A state represents an item, tool, location, environmental condition, player condition, structure, entity event, or abstract group. Directed typed edges specify how a task relates to these states. The resulting graph is\par
\begin{equation}
\label{eq:app-graph}
\begin{aligned}
G&=(\mathcal{T},\mathcal{S},\mathcal{E}),\\
\mathcal{E}&=\mathcal{E}_{\mathrm{req}}\cup\mathcal{E}_{\mathrm{prod}}\cup\mathcal{E}_{\mathrm{cons}}\cup\mathcal{E}_{\mathrm{enable}}\cup\mathcal{E}_{\mathrm{abs}}.
\end{aligned}
\end{equation}
\Needspace{5\baselineskip}
\refstepcounter{apptable}\label{tab:app-b2}
{\small Table \theapptable. Edge types in the released catalog. Counts describe graph definitions, before expansion into repeated stage occurrences.\par}\nopagebreak\smallskip
{\fontsize{9}{10.7}\selectfont\setlength{\tabcolsep}{2pt}\renewcommand{\arraystretch}{1.10}
\noindent\begin{minipage}{\linewidth}\begin{tabular}{>{\raggedright\arraybackslash}p{83.34pt}>{\raggedright\arraybackslash}p{151.54pt}>{\raggedright\arraybackslash}p{60.61pt}>{\raggedright\arraybackslash}p{83.34pt}}
\toprule
\textbf{Relation} & \textbf{Direction} & \textbf{Count} & \textbf{Meaning}\\\midrule
\texttt{requires} & State to task & 117 & Hard prerequisite for ordered composition\\
\texttt{produces} & Task to state & 81 & Declared output of a completed task\\
\texttt{consumes} & Task to state & 28 & Resource used by the task\\
\texttt{enables} & State to task & 15 & Soft capability relation for probe design\\
\texttt{abstracts\_\allowbreak{}to} & Node to group & 128 & Semantic grouping across concrete tasks and states\\
\bottomrule\end{tabular}\end{minipage}\par}
\smallskip
The relations in Eq.~\ref{eq:app-graph} are listed in Table~\ref{tab:app-b2}. The first three define the execution paths used in weighted graph progress. Soft enabling and abstraction edges organize capability relationships without contributing to this execution score. This makes the graph useful at two levels: concrete dependencies support composition, while abstract groups organize transfer questions across items or recipes.\par
\par\smallskip\noindent\begin{minipage}{\linewidth}
\centering\includegraphics[width=\linewidth]{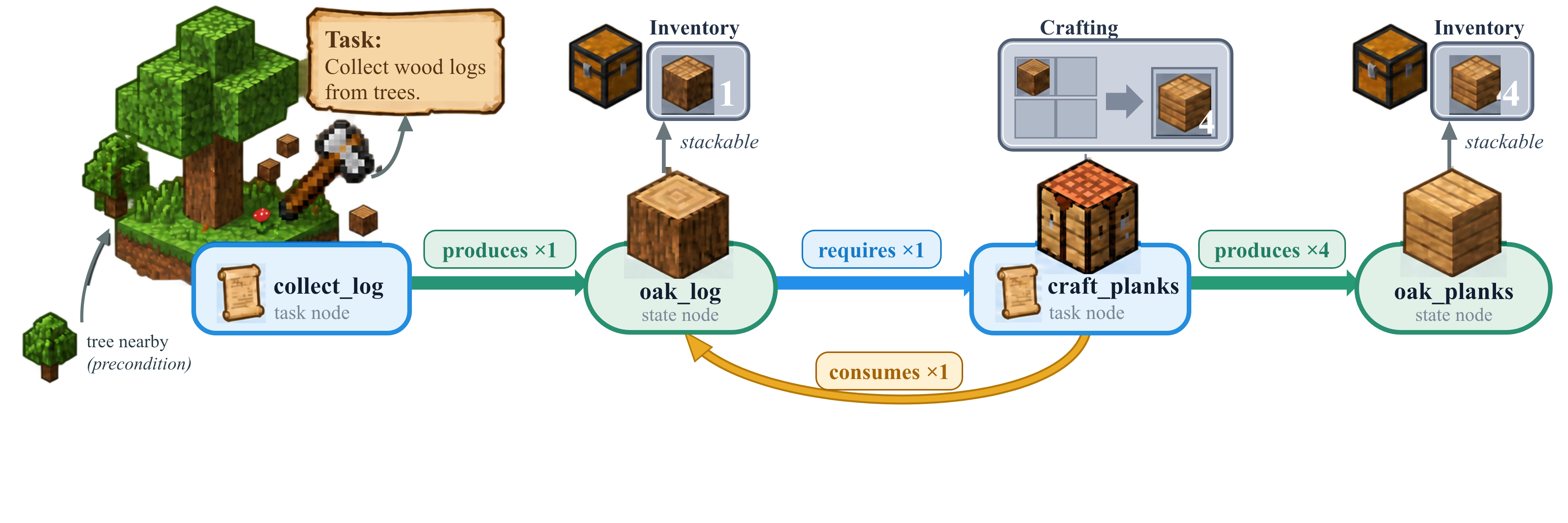}\par\smallskip
\refstepcounter{appfigure}\label{fig:app-b1}
{\small\raggedright Figure \theappfigure. A catalog fragment links log collection to plank production. Task nodes and state nodes have distinct shapes; edge labels retain their direction and quantity. The graph describes a declared dependency structure. Runtime observations and task predicates determine which execution stages actually pass.\par}
\end{minipage}\par\smallskip
\Needspace{5\baselineskip}
\subsection{Ordered chain compilation}
\label{app:tasks-3}
A chain stores an ordered task sequence, initial symbolic states, an environment profile, and a profile policy. The compiler traverses the sequence and extracts five paths for every occurrence: hard prerequisites, produced resources, consumed resources, soft enabling relations, and abstraction relations. Repeated task identifiers become distinct compiled stages, so their positions and local context remain available to the evaluator.\par
\Needspace{7\baselineskip}
The default compiler checks prerequisite presence in the evolving symbolic state set. An optional strict-quantity mode also checks declared resource counts before consuming and producing them. Let $q_j(s)$ be the symbolic availability of state $s$ before stage $j$. In the strict mode, a stage requires sufficient declared counts, after which its resource bookkeeping follows\par
\begin{equation}
\label{eq:app-quantity}
q_{j+1}(s)=q_j(s)-c_j(s)+p_j(s),
\end{equation}
where $c_j$ and $p_j$ are the stage's consumption and production counts. Initial symbolic states seed this calculation; they do not grant physical inventory. The simulator independently realizes movement, mining, crafting, and resource use.\par
This separation is useful when constructing a benchmark: a chain can be checked for intended ordering before a game is launched, and its execution can then be evaluated from observations. Equation~\ref{eq:app-quantity} describes symbolic bookkeeping only. The compiler\'s output also gives each stage a fixed metric context, with graph weights fixed by the planned chain before execution.\par
\Needspace{5\baselineskip}
\subsection{Composition patterns}
\label{app:tasks-4}
Resource progression chains connect acquisition to increasingly demanding use. For example, the ore progression chain moves through wood preparation, basic tools, stone tools, ore collection, smelting, and higher-tier mining. Cross-domain chains join heterogeneous endpoints: locating coal, obtaining fuel, creating illumination, building shelter, and remaining alive, or preparing a weapon before a combat event. Transfer chains substitute wood species, tool recipes, or item roles while retaining comparable dependency patterns.\par
The task suite also includes short probes. These isolate resource collection or limited crafting sequences, helping distinguish an early execution bottleneck from a long-horizon planning requirement. A short probe and a long chain answer different questions, even when they contain the same task identifier. Their sequence, start state, budget, and profile policy accompany their scores.\par
A completed possession task establishes its predicate at that stage. If the same item remains in inventory, a later occurrence with the same predicate may pass immediately. The ordered stage record makes this behavior explicit: initial-state passes have zero action steps, while newly acquired items appear in the observed state transition. Composition is consequently described through stage outcomes and state continuity, with experience transfer assessed by the reset protocol in Appendix~\ref{app:protocol}.\par
\Needspace{5\baselineskip}
\subsection{Extending the catalog}
\label{app:tasks-5}
A new task can be added by defining its instruction and measurable endpoint, identifying the required observation channels, selecting an environment profile, and attaching its graph relations. A new chain then specifies an ordered list of existing or new task identifiers. Compilation checks the declared dependency order, and an execution check verifies that the configured environment exposes the evidence required by its predicates.\par
The reusable components are the task contract, graph relations, profile registry, and runner interface. A method's planner, retrieval rule, or memory format remains outside the task definition. This allows a future agent to implement a different experience representation while being evaluated against the same target objectives and evidence rules.\par

\clearpage
\section{Success Verification and Execution Metrics}
\label{app:verification}
\Needspace{5\baselineskip}
\subsection{Predicate-based task evaluation}
\label{app:verification-1}
MC-EvoBench evaluates success from environment observations. A task contains a conjunction of predicates, with optional disjunctive subexpressions. Inventory thresholds check resource or item possession; equipment predicates accept an item in a specified slot; counters capture events such as an entity defeat; snapshot comparisons read a named observation field. These primitives cover both simple item goals and objectives that combine player state with local environmental conditions.\par
\Needspace{5\baselineskip}
\refstepcounter{apptable}\label{tab:app-c1}
{\small Table \theapptable. Supported predicate families and their measurement roles.\par}\nopagebreak\smallskip
{\fontsize{9}{10.7}\selectfont\setlength{\tabcolsep}{2pt}\renewcommand{\arraystretch}{1.10}
\noindent\begin{minipage}{\linewidth}\begin{tabular}{>{\raggedright\arraybackslash}p{88.26pt}>{\raggedright\arraybackslash}p{134.32pt}>{\raggedright\arraybackslash}p{161.18pt}}
\toprule
\textbf{Predicate family} & \textbf{Accepted evidence} & \textbf{Example}\\\midrule
Inventory threshold & Named item count at or above a threshold & At least four oak planks\\
Equipment equality & Normalized item in the specified equipment slot & Stone sword in the main hand\\
Counter threshold & Observed event count & At least one zombie defeated\\
Player-stat threshold & Named player measurement & Positive life or health\\
Snapshot comparison & Named path, comparison operator, and value & Observed coal-ore indicator at least one\\
Flag membership & Flag produced by the configured event logic & A mission state supported by its observation rule\\
Disjunction & At least one successful child predicate & Item in inventory or equipped\\
\bottomrule\end{tabular}\end{minipage}\par}
\smallskip
A snapshot-path comparison fails when its required path is absent. Inventory and counter thresholds use zero for a missing count. The diagnostic record stores the predicate and its observed value, making the distinction between a satisfied condition and unavailable supporting evidence inspectable. Item possession is measured directly from environment observations.\par
The evaluator checks the initial snapshot before requesting an action. If all predicates already hold, the stage passes with zero steps. Otherwise, it checks after each environment transition until success, a mission failure, environment termination, or budget exhaustion. A stage limit is the smaller of its catalog budget and any configured cap. The normal chain runner stops at the first non-passing stage, preserving the distinction between the observed prefix and unattempted suffix.\par
\Needspace{5\baselineskip}
\subsection{Mission events}
\label{app:verification-2}
Four mission-event definitions extend the basic predicates with observation-derived conditions. Their triggers combine local perception, lighting, elapsed time, and entity events. Event evaluation can attach a normalized flag or counter to the snapshot, after which the ordinary task verifier reads the corresponding evidence. This approach keeps event semantics shared between the runner and the task specification.\par
\Needspace{5\baselineskip}
\refstepcounter{apptable}\label{tab:app-c2}
{\small Table \theapptable. Mission-event rules. Thresholds are taken from the catalog. The shelter rule is a local structure-and-light criterion; the night rule measures a defined survival interval.\par}\nopagebreak\smallskip
{\fontsize{9}{10.7}\selectfont\setlength{\tabcolsep}{2pt}\renewcommand{\arraystretch}{1.10}
\noindent\begin{minipage}{\linewidth}\begin{tabular}{>{\raggedright\arraybackslash}p{88.26pt}>{\raggedright\arraybackslash}p{134.32pt}>{\raggedright\arraybackslash}p{161.18pt}}
\toprule
\textbf{Event} & \textbf{Trigger} & \textbf{Endpoint}\\\midrule
Coal source observed & Coal ore appears in the ray channel or local block grid & Observed resource discovery\\
Lit shelter observed & At least one local torch; at least 60 non-air blocks; effective light at least 7; sky light at most 7 or sky blocked & Local sheltered and lit condition\\
Night interval survived & Player alive; elapsed time at least 1,800 ticks; final time in the night window from 12,000 to 24,000 & Survival over the configured interval\\
Zombie defeat observed & Normalized zombie counter or raw zombie-kill count at least one & Observed combat event\\
\bottomrule\end{tabular}\end{minipage}\par}
\smallskip
For \texttt{survive\_\allowbreak{}first\_\allowbreak{}night}, the runner also derives \texttt{night\_\allowbreak{}survived} from player survival and elapsed time reaching the effective stage limit. This additional path uses the total-time difference when available; its world-time fallback also requires a night endpoint. Thus the effective interval follows a configured stage cap, while Table~\ref{tab:app-c2} specifies the catalog event rule. Survival, shelter, and combat events include a player-death failure condition. Event records preserve the evidence that caused a terminal decision, so later analysis can connect a stage outcome to the underlying world measurement.\par
\Needspace{5\baselineskip}
\subsection{Stage records and evidence organization}
\label{app:verification-3}
Each stage record identifies the task occurrence, terminal status, action-step count, and failed predicates. Its diagnostics include start and final snapshots, declared observation sources, actual predicate values, mission evidence, and the compiled graph paths. The chain report retains the planned stage count and planned graph weight even when execution stops early.\par
\Needspace{5\baselineskip}
\refstepcounter{apptable}\label{tab:app-c3}
{\small Table \theapptable. Terminal statuses and their interpretation in chain aggregation.\par}\nopagebreak\smallskip
{\fontsize{9}{10.7}\selectfont\setlength{\tabcolsep}{2pt}\renewcommand{\arraystretch}{1.10}
\noindent\begin{minipage}{\linewidth}\begin{tabular}{>{\raggedright\arraybackslash}p{88.26pt}>{\raggedright\arraybackslash}p{134.32pt}>{\raggedright\arraybackslash}p{161.18pt}}
\toprule
\textbf{Status} & \textbf{Interpretation} & \textbf{Contribution to passed-stage numerator}\\\midrule
\texttt{passed} & All success predicates hold & One\\
\texttt{failed} & A configured mission failure is observed & Zero\\
\texttt{terminated} & The environment ends before success & Zero\\
\texttt{timeout} & The effective stage limit is exhausted & Zero\\
\texttt{budget\_\allowbreak{}exhausted} & A smaller remaining chain/source budget is exhausted & Zero\\
Unattempted suffix & Earlier stopping prevents execution & Zero, with missing execution recorded separately\\
\bottomrule\end{tabular}\end{minipage}\par}
\smallskip
This record supports analysis at several resolutions. A chain score summarizes completed requirements; stage records locate its execution breakpoint; start and final states describe resource changes; and optional action traces support a finer behavioral account. These resolutions share the same task identity and sequence position.\par
\Needspace{5\baselineskip}
\subsection{Task success rate and weighted graph progress}
\label{app:verification-4}
\Needspace{7\baselineskip}
For chain $c$, let $L_c$ be its number of planned stage occurrences and $y_{cj}$ indicate whether occurrence $j$ passes. Task success rate is\par
\begin{equation}
\label{eq:app-tsr}
\mathrm{TSR}=\frac{1}{L_c}\sum_{j=1}^{L_c}y_{cj}.
\end{equation}
In Eq.~\ref{eq:app-tsr}, unattempted stages remain in the planned denominator. At split level, the aggregate pools passed occurrences and planned occurrences across the included chains. This weights every planned occurrence equally, while completed-chain counts offer a complementary measure of end-to-end execution.\par
\Needspace{7\baselineskip}
For chain $c$, let $E_{\mathrm{exec},c}$ be its strict execution edges, formed by the \texttt{requires}, \texttt{produces}, and \texttt{consumes} relations. Let $w_e$ be the edge weight and $q_e$ indicate whether the corresponding verified stage contributes. Edge weights default to one when unspecified. Weighted graph progress is\par
\begin{equation}
\label{eq:app-wgp}
\mathrm{WGP}=\frac{\sum_{e\in E_{\mathrm{exec},c}}w_e q_e}{\sum_{e\in E_{\mathrm{exec},c}}w_e}.
\end{equation}
In Eq.~\ref{eq:app-wgp}, an edge is credited when its corresponding verified stage contributes. Shared resources can appear in several stage paths, and each occurrence contributes its local edge weight. Split WGP pools completed and planned edge weights across chains. A chain with no planned execution-edge weight falls back to TSR. Together, TSR and WGP summarize stage completion and its execution structure using fixed planned denominators.\par
\Needspace{5\baselineskip}
\subsection{Milestone and efficiency diagnostics}
\label{app:verification-5}
\Needspace{7\baselineskip}
Verified milestone progress (VMP) describes observed acquisition of graph-linked items or tools. For each reported stage, the metric extracts item/tool states from its execution paths and assigns each state the largest included edge-count threshold. A state is eligible if the first recorded snapshot is below its threshold; it is credited if a later recorded stage-start or stage-final snapshot reaches the threshold. The score is\par
\begin{equation}
\label{eq:app-vmp}
\mathrm{VMP}=\frac{\text{eligible states reaching their thresholds}}{\text{eligible states}},
\end{equation}
Equation~\ref{eq:app-vmp} returns zero when the eligible set is empty. The associated eligible count is part of its interpretation. Because this set follows the reported prefix, VMP is used as a local acquisition diagnostic alongside the fixed-denominator TSR and WGP endpoints. A task passes when its success predicates hold.\par
Efficiency summaries include environment steps, stage budgets, steps per passed stage, and observed deaths. Resource-use diagnostics identify consumed items whose inventory decreases during a non-passing stage. This provides a traceable signal of material expenditure associated with unsuccessful execution. Recovery events require an explicit adapter record and can then be aligned with stage transitions.\par
\Needspace{5\baselineskip}
\subsection{Verification examples}
\label{app:verification-6}
The task \texttt{craft\_\allowbreak{}planks} requires at least four oak planks. The objective is therefore satisfied by the corresponding inventory count. For \texttt{craft\_\allowbreak{}stone\_\allowbreak{}sword}, the catalog accepts a stone sword either in inventory or in the main hand, matching the game's distinction between carrying and equipping a tool. The coal-discovery task uses observed coal in a ray or grid channel; the subsequent coal-collection task checks the collected item.\par
These examples illustrate why the task contract separates an instruction, its observation sources, and its terminal predicate. The instruction gives the agent a goal, the observation contract defines the evidence channels, and the predicate fixes the endpoint. The same decomposition supports independently checking an item goal, a temporal objective, or a cross-domain chain through explicit predicates.\par
\clearpage
\section{Evaluation protocol}
\label{app:protocol}
\Needspace{5\baselineskip}
\subsection{Experience, artifact, and target evaluation}
\label{app:protocol-1}
MC-EvoBench separates capability measurement from the experience that may change an agent. A paired evaluation first measures a target with the base agent, then measures the same target with an artifact produced through a specified source curriculum. The artifact may contain updated parameters, memory, a skill library, or another declared persistent representation. Its identity and provenance define the agent state being evaluated. This interface accommodates different adaptation mechanisms while keeping the target task and its verifier fixed.\par
At the source-to-target boundary, the target world and inventory are reset. Only the frozen method-produced artifact crosses this boundary. Consequently, tools, blocks, completed constructions, and resource inventories accumulated during source interaction do not supply target progress. In the supplied memory configuration, source interaction uses \texttt{read\_\allowbreak{}write} memory and target evaluation uses \texttt{read\_\allowbreak{}only} memory. Each target role constructs a fresh agent and environment; persistent source experience is loaded from its declared artifact.\par
Within a compiled chain, stages execute in order and preserve the environment state permitted by the chain profile. The runner stops at the first non-passing stage. The transfer probes use one chain-level profile throughout an episode. Source episodes can restart to complete their interaction budget, as described in Appendix~\ref{app:protocol-3}. These distinct boundaries preserve sequential task dependencies within an episode and isolate learned experience across evaluations.\par
\par\smallskip\noindent\begin{minipage}{\linewidth}
\centering\includegraphics[width=\linewidth]{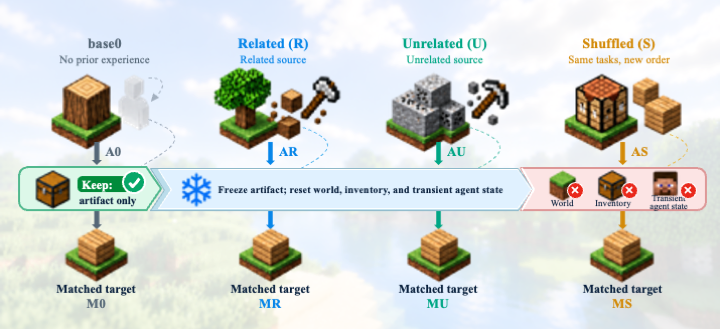}\par\smallskip
\refstepcounter{appfigure}\label{fig:app-d1}
{\small\raggedright Figure \theappfigure. Each source condition produces its own artifact from the same base agent. Target evaluations share task, world, budget, and seed settings. At the boundary, the world, inventory, and transient agent state are reset while the declared persistent artifact is retained.\par}
\end{minipage}\par\smallskip
\Needspace{5\baselineskip}
\subsection{Paired target roles and controls}
\label{app:protocol-2}
Each probe and replicate defines three source roles and four target roles. All four target roles use the same target task sequence, verifier, environment profile, world seed, stage limits, and agent seed. They differ in the artifact supplied to the agent.\par
\Needspace{5\baselineskip}
\refstepcounter{apptable}\label{tab:app-d1}
{\small Table \theapptable. Target roles and the question addressed by each source condition.\par}\nopagebreak\smallskip
{\fontsize{9}{10.7}\selectfont\setlength{\tabcolsep}{2pt}\renewcommand{\arraystretch}{1.10}
\noindent\begin{minipage}{\linewidth}\begin{tabular}{>{\raggedright\arraybackslash}p{88.26pt}>{\raggedright\arraybackslash}p{134.32pt}>{\raggedright\arraybackslash}p{161.18pt}}
\toprule
\textbf{Target role} & \textbf{Artifact supplied} & \textbf{Comparison supported}\\\midrule
No prior & Base artifact without the evaluated source experience & Initial target capability\\
Related & Artifact after graph-related source tasks & Transfer from relevant experience\\
Unrelated & Artifact after unrelated source tasks & Effect of generic extra experience\\
Shuffled & Artifact after the related task multiset in reversed order & Sensitivity to source ordering\\
\bottomrule\end{tabular}\end{minipage}\par}
\smallskip
The unrelated and shuffled source conditions receive the same family-level environment-step budget as the related condition. The shuffled source uses the identical task multiset, including repeated tasks. Its probe explicitly allows compilation with unmet prerequisites, making the reversed sequence executable without adding inventory or changing success predicates. Normal runtime stopping rules still apply, so the resulting experience can include repeated unsuccessful episodes.\par
The target input contract provides the natural-language instruction and agent observation, together with the declared artifact. It reserves graph-relation manifests, verifier predicates, verifier diagnostics, and hidden resource coordinates for evaluation. The built-in method inputs are detailed in Appendix~\ref{app:agents}. Public resource placement defines the controlled task instance. It does not supply target inventory items, forced success evidence, private route hints, or scripted actions.\par
\Needspace{5\baselineskip}
\subsection{Frozen transfer probes and source exposure}
\label{app:protocol-3}
The protocol provides three shorter graph-chain families assembled from catalog tasks. Shorter targets make transfer at intermediate capabilities observable even when a long progression chain stops at an early prerequisite. The plan freezes task sequences, resource profiles, source--target relations, and budgets before adaptive paired evaluation.\par
\Needspace{5\baselineskip}
\refstepcounter{apptable}\label{tab:app-d2}
{\small Table \theapptable. Transfer probe families. Source exposure is matched within each row; the target sequence is held fixed across artifact conditions.\par}\nopagebreak\smallskip
{\fontsize{9}{10.7}\selectfont\setlength{\tabcolsep}{2pt}\renewcommand{\arraystretch}{1.10}
\noindent\begin{minipage}{\linewidth}\begin{tabular}{>{\raggedright\arraybackslash}p{83.34pt}>{\raggedright\arraybackslash}p{151.54pt}>{\raggedright\arraybackslash}p{60.61pt}>{\raggedright\arraybackslash}p{83.34pt}}
\toprule
\textbf{Probe family} & \textbf{Related source $\rightarrow$ target variation} & \textbf{Stages per sequence} & \textbf{Source steps per condition}\\\midrule
Wood species & Birch logs and planks $\rightarrow$ acacia logs and planks & 2 & 900\\
Wooden tools & Wood preparation and wooden axe $\rightarrow$ wooden shovel & 5 & 2,250\\
Stone tools & Wood/stone preparation and stone pickaxe $\rightarrow$ stone shovel & 8 & 3,600\\
\bottomrule\end{tabular}\end{minipage}\par}
\smallskip
The wooden-tool pair shares the log, plank, stick, and crafting-table preparation sequence. The stone-tool pair additionally shares wooden-pickaxe preparation and two cobblestone-mining stages. Unrelated source sequences contain the same number of stages and use collection tasks outside the selected production chain. The resource profiles provide local world resources while retaining the need to acquire and transform them through agent interaction.\par
The probe configuration imposes a 450-step ceiling on each stage, retaining a lower catalog limit if one exists. The fixed-exposure source runner is enabled by \texttt{experiment.source\_\allowbreak{}interaction\_\allowbreak{}budget\_\allowbreak{}steps}. Source exposure is measured as actual environment interactions. If a source episode ends before its family budget is consumed, the agent receives normal terminal feedback, the source environment resets, and interaction resumes with the persistent method artifact. The last episode is truncated at the remaining budget. Thus early completion or failure changes the number of episodes without reducing the prescribed source exposure. The source report records every episode and the total interaction count; a final paired comparison requires all three source conditions to consume the common budget.\par
\Needspace{5\baselineskip}
\subsection{Replicates and reproducibility}
\label{app:protocol-4}
The headline paired protocol uses five final agent seeds, 5100--5104. Separate calibration checks may use at least three direct seeds, but they are not part of the 105 role configurations or the headline results. The replicate axis is agent randomness on a fixed controlled world: the public world seed is 31. Fixing the world preserves the declared spawn placement and local resource layout across conditions. The resulting paired estimates characterize variation across agent seeds on these controlled instances.\par
Run labels such as \texttt{seed0} identify records; numerical \texttt{world\_\allowbreak{}seed} and \texttt{agent\_\allowbreak{}seed} fields determine randomization. The runner applies the world seed to the environment and seeds Python, NumPy, and PyTorch before constructing the policy. Reports retain both seeds and their provenance. Every role in a paired replicate uses the same numerical seeds, and the same replicate list is used across conditions.\par
Before rollout, the launcher checks the declared resource profile through an environment reset. Artifact manifests record identities and paths, with SHA-256 fingerprints available for files and directory bundles. Together, the frozen plan, explicit seeds, artifact identity, and runtime configuration specify a reproducible comparison.\par
\Needspace{5\baselineskip}
\subsection{Transfer contrasts}
\label{app:protocol-5}
\Needspace{7\baselineskip}
TSR, WGP, and VMP follow the definitions in Appendix~\ref{app:verification}. Let $M_0,M_R,M_U,M_S$ denote a target score with no prior, related, unrelated, and shuffled artifacts, respectively. Define each source condition's change from the no-prior target as\par
\begin{equation}
\label{eq:app-contrast}
\Delta_c=M_c-M_0,\qquad c\in\{R,U,S\}.
\end{equation}
The paired differences in Eq.~\ref{eq:app-contrast} are interpreted across the three direct metrics. A related-source change is compared with both the no-prior baseline and the matched unrelated and shuffled controls; these are contrasts over TSR, WGP, and VMP rather than additional benchmark metrics. Target TSR is the primary channel, while WGP and VMP provide complementary completion and acquisition diagnostics.\par
\Needspace{5\baselineskip}
\subsection{Statistical and diagnostic reporting}
\label{app:protocol-6}
\Needspace{7\baselineskip}
Compute contrasts within each matched replicate before summarizing across seeds. Report the four target scores and paired differences for each direct metric, with their means and sample standard deviations. The aggregation helper also computes two-sided 95\% Student-t intervals on paired seed contrasts:\par
\begin{equation}
\label{eq:app-interval}
\bar d\ \pm\ t_{0.975,n-1}\frac{s_d}{\sqrt n}.
\end{equation}
For the interval in Eq.~\ref{eq:app-interval}, $n$ is the number of valid pairs and $s_d$ is the sample standard deviation of paired differences. A beneficial-transfer interpretation requires a positive related-source change relative to the no-prior baseline and matched controls, supported by paired uncertainty. Metric configurations are also characterized as floor-limited, ceiling-limited, or interior. The accompanying progress channels localize capability acquisition across these score regimes.\par
Report environment steps alongside source episode counts, outcomes, memory updates, planner calls, and token usage. Equal environment exposure can produce different amounts of planning and updating because episodes terminate at different points. These diagnostics characterize the resources used to create each artifact and support computation-aware comparisons. Per-seed target scores and source exposure remain available alongside aggregate statistics, allowing transfer and efficiency to be assessed together.\par
\Needspace{8\baselineskip}
\subsection{Per-seed transfer records}
\label{app:protocol-7}
Table~\ref{tab:app-d4} lists the per-seed target TSR values behind Figure~\ref{fig:stage-seed}b--c and Table~\ref{tab:transfer-contrasts}, computed within each matched replicate as described in Appendix~\ref{app:protocol-6}.\par
\Needspace{12\baselineskip}
\refstepcounter{apptable}\label{tab:app-d4}
{\small Table \theapptable. Per-seed target TSR under the four target conditions: no prior (N), related (R), unrelated (U), and shuffled (S). $\Delta R=R-N$ and $\mathrm{GATG}=R-\frac{1}{2}(U+S)$ are computed within each seed; values are rounded to three decimals.\par}\nopagebreak\smallskip
{\fontsize{9}{10.7}\selectfont\renewcommand{\arraystretch}{1.05}
\noindent\begin{tabular*}{\linewidth}{@{\extracolsep{\fill}}llrrrrrr@{}}
\toprule
\textbf{Probe family} & \textbf{Seed} & \textbf{N} & \textbf{R} & \textbf{U} & \textbf{S} & $\boldsymbol{\Delta R}$ & \textbf{GATG}\\\midrule
Wood species & \texttt{seed0} & 0.000 & 0.000 & 0.000 & 1.000 & 0.000 & $-0.500$\\
 & \texttt{seed1} & 0.500 & 1.000 & 0.000 & 0.000 & 0.500 & 1.000\\
 & \texttt{seed2} & 0.500 & 0.000 & 0.000 & 1.000 & $-0.500$ & $-0.500$\\
 & \texttt{seed3} & 1.000 & 0.000 & 0.000 & 0.000 & $-1.000$ & 0.000\\
 & \texttt{seed4} & 0.000 & 1.000 & 1.000 & 0.500 & 1.000 & 0.250\\\midrule
Wooden tool & \texttt{seed0} & 0.200 & 0.200 & 0.200 & 0.200 & 0.000 & 0.000\\
 & \texttt{seed1} & 0.200 & 0.000 & 0.000 & 0.000 & $-0.200$ & 0.000\\
 & \texttt{seed2} & 0.800 & 0.400 & 0.400 & 0.200 & $-0.400$ & 0.100\\
 & \texttt{seed3} & 0.200 & 0.400 & 0.200 & 0.400 & 0.200 & 0.100\\
 & \texttt{seed4} & 0.200 & 0.000 & 0.000 & 0.400 & $-0.200$ & $-0.200$\\\midrule
Stone tool & \texttt{seed0} & 0.125 & 0.125 & 0.000 & 0.000 & 0.000 & 0.125\\
 & \texttt{seed1} & 0.125 & 0.000 & 0.000 & 0.125 & $-0.125$ & $-0.063$\\
 & \texttt{seed2} & 0.125 & 0.000 & 0.000 & 0.000 & $-0.125$ & 0.000\\
 & \texttt{seed3} & 0.125 & 0.125 & 0.125 & 0.125 & 0.000 & 0.000\\
 & \texttt{seed4} & 0.250 & 0.000 & 0.000 & 0.125 & $-0.250$ & $-0.063$\\
\bottomrule\end{tabular*}\par}
\smallskip
\clearpage
\section{Agent implementations}
\label{app:agents}
\Needspace{5\baselineskip}
\subsection{Observation and action interfaces}
\label{app:agents-1}
MC-EvoBench connects policies to a common chain runner through a policy factory, a reset operation, and an action function. At each interaction, the adapter supplies the task instruction and current observation and receives a MineRL action mapping. The default prompt template is the instruction itself. Optional callbacks provide stage status and the number of executed steps, allowing methods with persistent memory to update after source experience. Table~\ref{tab:app-e1} specifies the implemented method variants and their effective inputs; these interfaces define the observation, action, and persistent-state conditions used in each comparison. The table documents supported adapters; the headline reference study reported in the main paper uses only the JARVIS-1 online extension, while the other entries do not contribute to the paired-transfer results.\par
\Needspace{5\baselineskip}
\refstepcounter{apptable}\label{tab:app-e1}
{\small Table \theapptable. Implemented agent interfaces. ``Persistent state'' identifies information retained by the bridge for later decisions.\par}\nopagebreak\smallskip
{\fontsize{9}{10.7}\selectfont\setlength{\tabcolsep}{2pt}\renewcommand{\arraystretch}{1.10}
\noindent\begin{minipage}{\linewidth}\begin{tabular}{>{\raggedright\arraybackslash}p{88.26pt}>{\raggedright\arraybackslash}p{134.32pt}>{\raggedright\arraybackslash}p{161.18pt}}
\toprule
\textbf{Variant} & \textbf{Effective decision input} & \textbf{Action and persistent state}\\\midrule
VPT & Raw MineRL observation; instruction unused & Released visual policy; recurrent controller state\\
STEVE-1 / JARVIS-1 controller & Observation and text embedding & Released STEVE-I actions; cached prompt embeddings\\
JARVIS online extension & Planner: inventory, equipment, position, retrieved episodes; controller: observation and subgoal & Local text planner with released controller; episode memory\\
ExpeL adaptation & Inventory, equipment, position/orientation, nearby blocks, line of sight, episodes and insights & LLM-generated action mapping; episodes, insights and weighted rules\\
Optimus-1 action bridge & Screenshot file and instruction & Released action model; direct controller execution\\
Optimus-1 planner bridge & Planner: screenshot and instruction; controller: screenshot and subgoal & Local visual planner with released action model; diagnostic trace\\
Optimus-3 action path & Image and task-conditioned embedding & Released MLLM, task router and action head\\
Voyager skill bridge & Fixed task-to-skill sequence and Mineflayer state & Released JavaScript skills and primitives; skill files\\
\bottomrule\end{tabular}\end{minipage}\par}
\smallskip
The STEVE-1 and JARVIS controller bridges encode text with MineCLIP and a learned prior, cache embeddings by prompt, and reset the conditional controller with guidance scale 6.0. VPT loads its network specification and policy-head settings from the model file; its sampling temperature is inherited from that file. Optimus-3 creates a task representation when the instruction changes, then combines it with successive frames.\par
\Needspace{5\baselineskip}
\subsection{Persistent experience and adaptation}
\label{app:agents-2}
The JARVIS online extension stores stage-level textual records containing the instruction, proposed plan, dispatched controller prompts, status, interaction count and visible-state summary. Retrieval uses token-set Jaccard similarity between the current instruction and each stored instruction, adds 0.15 for a successful record, and breaks score ties in favour of newer records. Up to four positive-scoring records condition planning. Memory retains the latest 256 records and is written atomically. The planner returns one to six short controller prompts; reflection produces a practical sentence capped at 500 characters. The reflection input distinguishes a proposed plan from prompts actually dispatched and labels the state summary as the last pre-action observation.\par
ExpeL uses the same episode-retrieval score and supplements episodes with retrieved textual insights and a weighted rule list. Insight retrieval uses lexical overlap; rule retrieval adds a weight contribution capped at 0.10. With the configured insight limit of eight, the planner may receive up to eight insights and eight rules. Reflection proposes ADD, EDIT, REMOVE and AGREE operations. New rules start with weight two; agreement and editing increment weight, while removal decrements it. The persistent artifact records these updates alongside episode actions. Both memory methods support read-write source interaction and read-only target evaluation; read-only stage callbacks suppress reflection and memory writes.\par
\Needspace{5\baselineskip}
\subsection{Planning and execution cadence}
\label{app:agents-3}
The JARVIS extension plans when a new stage instruction becomes active. It dispatches successive subgoals for fixed 120-step intervals and retains the final subgoal for any remaining stage steps. Stage completion invalidates the active instruction so repeated task text can trigger a new plan. Optimus-1's visual planner uses the same interval rule and a JSON subgoal format. Its input is a JPEG thumbnail bounded by 384 $\times$ 384 pixels at quality 85. The parser filters subgoals containing ``open,'' ``place,'' or ``access'' followed by a space, rewrites punch/collect/gather to chop, and retains up to six subgoals. ExpeL instead requests a low-level action mapping and can hold that action across a configured number of environment steps. The 19-chain configuration sets this hold to 32, while the factory default is one. Its prompt also supplies explicit line-of-sight guidance for approaching and mining visible resources.\par
The Voyager skill bridge follows a separate execution protocol. The runner loads released JavaScript skills and control primitives, resets through the Mineflayer bridge with empty inventory and equipment, and invokes a fixed skill sequence. A configured starting position is applied during setup. The runner checks the returned inventory for its designated target item. Its step count denotes a bridge request, and the implementation invokes no LLM; this execution unit is recorded separately from MineRL interaction steps.\par
\Needspace{5\baselineskip}
\subsection{Configured parameters and replication}
\label{app:agents-4}
Table~\ref{tab:app-e2} reports configuration values and the defaults used where configurations omit a field. Planner model identifiers are recorded literally because the standard and relay matrices select different providers. Planner calls leave temperature unspecified. Token limits bound generated output, and timeout or retry settings describe request handling rather than additional environment interaction.\par
\Needspace{5\baselineskip}
\refstepcounter{apptable}\label{tab:app-e2}
{\small Table \theapptable. Configuration values and implementation defaults. J = JARVIS online extension; O = Optimus-1 planner bridge; E = ExpeL adaptation.\par}\nopagebreak\smallskip
{\fontsize{9}{10.7}\selectfont\setlength{\tabcolsep}{2pt}\renewcommand{\arraystretch}{1.10}
\noindent\begin{minipage}{\linewidth}\begin{tabular}{>{\raggedright\arraybackslash}p{128.26pt}>{\raggedright\arraybackslash}p{260.42pt}}
\toprule
\textbf{Component} & \textbf{Configured value or applicable default}\\\midrule
Text-conditioned controllers & VPT \texttt{2x.model}; STEVE-1 weights and prior; guidance 6.0 for STEVE-1/JARVIS\\
J standard / relay planner & \texttt{gpt-\allowbreak{}5.4} / \texttt{gpt-\allowbreak{}5.6-\allowbreak{}sol}; relay uses Responses with high reasoning effort\\
O / E planner & \texttt{\textasciitilde{}openai/\allowbreak{}gpt-\allowbreak{}luna-\allowbreak{}latest} / \texttt{gpt-\allowbreak{}5.6-\allowbreak{}sol}; E uses Responses, high effort\\
Output limit; timeout & J: 4,096 tokens, 60 s; O: 4,096, 90 s; E: 2,048, 60 s\\
Memory and cadence & 256 records; 4 retrieved episodes; J/O horizon 120; E insight/rule limits 8 each, action hold 32\\
\bottomrule\end{tabular}\end{minipage}\par}
\smallskip
The standard and relay configurations instantiate the same benchmark roles through different planner interfaces. A reproducible method specification records the selected matrix, policy checkpoint, observation channels, action cadence, planner identity, and memory mode. Replicate seeds and source budgets follow Appendix~\ref{app:protocol}. These fields allow the effect of the experience mechanism to be interpreted alongside its controller and computational settings.\par
\clearpage
\section{Complete Benchmark Suite}
\label{app:suite}
\Needspace{5\baselineskip}
\subsection{Task inventory}
\label{app:suite-1}
Tables~\ref{tab:app-f1}--\ref{tab:app-f3} enumerate the 80 task definitions once. The code T01--T80 is a compact cross-reference used only in this appendix; the adjacent identifier is the executable catalog key. ``Steps'' is the catalog stage budget, before an experiment applies a smaller cap. Profile codes refer to Table~\ref{tab:app-a2}. Success conditions specify the required observed endpoint; event abbreviations refer to the exact rules in Appendix~\ref{app:verification}.\par
The supplementary CSV preserves each original natural-language instruction and the full JSON success specification. This keeps the printed table compact while providing an exact machine-readable mapping between the manuscript, task identifiers, and evaluator predicates. In a chain, each occurrence retains its task's definition and receives the profile selected by the chain policy.\par
\Needspace{5\baselineskip}
\refstepcounter{apptable}\label{tab:app-f1}
{\small Table \theapptable. Core Progression: 33 task definitions.\par}\nopagebreak\smallskip
{\fontsize{9}{10.7}\selectfont\setlength{\tabcolsep}{2pt}\renewcommand{\arraystretch}{1.10}
\begin{longtable}{>{\raggedright\arraybackslash}p{159.11pt}>{\raggedright\arraybackslash}p{143.96pt}>{\raggedright\arraybackslash}p{30.31pt}>{\raggedright\arraybackslash}p{45.46pt}}
\toprule
\textbf{Code / catalog task} & \textbf{Success condition} & \textbf{Steps} & \textbf{Profile}\\\midrule
\endfirsthead
\toprule
\textbf{Code / catalog task} & \textbf{Success condition} & \textbf{Steps} & \textbf{Profile}\\\midrule
\endhead
\bottomrule\endfoot
T01 \,\textperiodcentered\, \texttt{scout\_\allowbreak{}coal\_\allowbreak{}source} & Coal visible in ray or local grid (\ref{app:verification-2}) & 500 & P05\\
T02 \,\textperiodcentered\, \texttt{collect\_\allowbreak{}log} & oak log $\geq$1 & 400 & P02\\
T03 \,\textperiodcentered\, \texttt{craft\_\allowbreak{}planks} & oak planks $\geq$4 & 600 & P02\\
T04 \,\textperiodcentered\, \texttt{craft\_\allowbreak{}table} & crafting table $\geq$1 & 700 & P02\\
T05 \,\textperiodcentered\, \texttt{craft\_\allowbreak{}wooden\_\allowbreak{}pickaxe} & wooden pickaxe $\geq$1 & 1000 & P02\\
T06 \,\textperiodcentered\, \texttt{collect\_\allowbreak{}coal} & coal $\geq$1 & 700 & P05\\
T07 \,\textperiodcentered\, \texttt{craft\_\allowbreak{}torch} & torch $\geq$1 & 220 & P05\\
T08 \,\textperiodcentered\, \texttt{build\_\allowbreak{}lit\_\allowbreak{}shelter} & Local structure and lighting rule (\ref{app:verification-2}) & 1200 & P04\\
T09 \,\textperiodcentered\, \texttt{survive\_\allowbreak{}first\_\allowbreak{}night} & Night-survival event (\ref{app:verification-2}) & 1800 & P03\\
T10 \,\textperiodcentered\, \texttt{mine\_\allowbreak{}cobblestone} & cobblestone $\geq$2 & 700 & P02\\
T11 \,\textperiodcentered\, \texttt{craft\_\allowbreak{}stone\_\allowbreak{}sword} & Stone sword in inventory $\geq$1 or in main hand & 280 & P02\\
T12 \,\textperiodcentered\, \texttt{defeat\_\allowbreak{}zombie} & Zombie-defeat counter $\geq$1 & 1200 & P03\\
T13 \,\textperiodcentered\, \texttt{craft\_\allowbreak{}sticks} & stick $\geq$4 & 300 & P02\\
T14 \,\textperiodcentered\, \texttt{craft\_\allowbreak{}furnace} & furnace $\geq$1 & 700 & P02\\
T15 \,\textperiodcentered\, \texttt{collect\_\allowbreak{}iron\_\allowbreak{}ore\_\allowbreak{}basic} & iron ore $\geq$1 & 1200 & P12\\
T16 \,\textperiodcentered\, \texttt{smelt\_\allowbreak{}iron\_\allowbreak{}ingot} & iron ingot $\geq$1 & 1200 & P12\\
T17 \,\textperiodcentered\, \texttt{collect\_\allowbreak{}string} & string $\geq$1 & 900 & P02\\
T18 \,\textperiodcentered\, \texttt{collect\_\allowbreak{}iron\_\allowbreak{}ore} & iron ore $\geq$1 & 1200 & P12\\
T19 \,\textperiodcentered\, \texttt{collect\_\allowbreak{}gold\_\allowbreak{}ore} & gold ore $\geq$1 & 1400 & P12\\
T20 \,\textperiodcentered\, \texttt{collect\_\allowbreak{}redstone} & redstone $\geq$1 & 1500 & P12\\
T21 \,\textperiodcentered\, \texttt{collect\_\allowbreak{}lapis\_\allowbreak{}lazuli} & lapis lazuli $\geq$1 & 1500 & P12\\
T22 \,\textperiodcentered\, \texttt{collect\_\allowbreak{}diamond} & diamond $\geq$1 & 2000 & P12\\
T23 \,\textperiodcentered\, \texttt{craft\_\allowbreak{}stone\_\allowbreak{}pickaxe} & stone pickaxe $\geq$1 & 800 & P02\\
T24 \,\textperiodcentered\, \texttt{smelt\_\allowbreak{}charcoal} & charcoal $\geq$1 & 900 & P02\\
T25 \,\textperiodcentered\, \texttt{smelt\_\allowbreak{}glass} & glass $\geq$1 & 900 & P02\\
T26 \,\textperiodcentered\, \texttt{smelt\_\allowbreak{}gold\_\allowbreak{}ingot} & gold ingot $\geq$1 & 1000 & P12\\
T27 \,\textperiodcentered\, \texttt{craft\_\allowbreak{}iron\_\allowbreak{}pickaxe} & iron pickaxe $\geq$1 & 1200 & P12\\
T28 \,\textperiodcentered\, \texttt{collect\_\allowbreak{}white\_\allowbreak{}wool} & white wool $\geq$1 & 700 & P02\\
T29 \,\textperiodcentered\, \texttt{craft\_\allowbreak{}white\_\allowbreak{}bed} & white bed $\geq$1 & 800 & P02\\
T30 \,\textperiodcentered\, \texttt{craft\_\allowbreak{}bow} & bow $\geq$1 & 800 & P02\\
T31 \,\textperiodcentered\, \texttt{craft\_\allowbreak{}shield} & shield $\geq$1 & 1000 & P02\\
T32 \,\textperiodcentered\, \texttt{craft\_\allowbreak{}bucket} & bucket $\geq$1 & 1000 & P02\\
T33 \,\textperiodcentered\, \texttt{collect\_\allowbreak{}water\_\allowbreak{}bucket} & water bucket $\geq$1 & 700 & P02\\
\end{longtable}}
\smallskip
\Needspace{5\baselineskip}
\refstepcounter{apptable}\label{tab:app-f2}
{\small Table \theapptable. Transfer and Schema Generalization: 31 task definitions.\par}\nopagebreak\smallskip
{\fontsize{9}{10.7}\selectfont\setlength{\tabcolsep}{2pt}\renewcommand{\arraystretch}{1.10}
\begin{longtable}{>{\raggedright\arraybackslash}p{159.11pt}>{\raggedright\arraybackslash}p{143.96pt}>{\raggedright\arraybackslash}p{30.31pt}>{\raggedright\arraybackslash}p{45.46pt}}
\toprule
\textbf{Code / catalog task} & \textbf{Success condition} & \textbf{Steps} & \textbf{Profile}\\\midrule
\endfirsthead
\toprule
\textbf{Code / catalog task} & \textbf{Success condition} & \textbf{Steps} & \textbf{Profile}\\\midrule
\endhead
\bottomrule\endfoot
T34 \,\textperiodcentered\, \texttt{craft\_\allowbreak{}chest} & chest $\geq$1 & 500 & P02\\
T35 \,\textperiodcentered\, \texttt{craft\_\allowbreak{}oak\_\allowbreak{}boat} & oak boat $\geq$1 & 600 & P02\\
T36 \,\textperiodcentered\, \texttt{craft\_\allowbreak{}wooden\_\allowbreak{}axe} & wooden axe $\geq$1 & 800 & P02\\
T37 \,\textperiodcentered\, \texttt{craft\_\allowbreak{}wooden\_\allowbreak{}sword} & wooden sword $\geq$1 & 600 & P02\\
T38 \,\textperiodcentered\, \texttt{craft\_\allowbreak{}wooden\_\allowbreak{}shovel} & wooden shovel $\geq$1 & 600 & P02\\
T39 \,\textperiodcentered\, \texttt{collect\_\allowbreak{}birch\_\allowbreak{}log} & birch log $\geq$1 & 450 & P02\\
T40 \,\textperiodcentered\, \texttt{craft\_\allowbreak{}birch\_\allowbreak{}planks} & birch planks $\geq$4 & 600 & P02\\
T41 \,\textperiodcentered\, \texttt{craft\_\allowbreak{}birch\_\allowbreak{}boat} & birch boat $\geq$1 & 700 & P02\\
T42 \,\textperiodcentered\, \texttt{collect\_\allowbreak{}spruce\_\allowbreak{}log} & spruce log $\geq$1 & 450 & P02\\
T43 \,\textperiodcentered\, \texttt{craft\_\allowbreak{}spruce\_\allowbreak{}planks} & spruce planks $\geq$4 & 600 & P02\\
T44 \,\textperiodcentered\, \texttt{craft\_\allowbreak{}spruce\_\allowbreak{}boat} & spruce boat $\geq$1 & 700 & P02\\
T45 \,\textperiodcentered\, \texttt{collect\_\allowbreak{}jungle\_\allowbreak{}log} & jungle log $\geq$1 & 450 & P02\\
T46 \,\textperiodcentered\, \texttt{craft\_\allowbreak{}jungle\_\allowbreak{}planks} & jungle planks $\geq$4 & 600 & P02\\
T47 \,\textperiodcentered\, \texttt{craft\_\allowbreak{}jungle\_\allowbreak{}boat} & jungle boat $\geq$1 & 700 & P02\\
T48 \,\textperiodcentered\, \texttt{collect\_\allowbreak{}acacia\_\allowbreak{}log} & acacia log $\geq$1 & 450 & P02\\
T49 \,\textperiodcentered\, \texttt{craft\_\allowbreak{}acacia\_\allowbreak{}planks} & acacia planks $\geq$4 & 600 & P02\\
T50 \,\textperiodcentered\, \texttt{craft\_\allowbreak{}acacia\_\allowbreak{}boat} & acacia boat $\geq$1 & 700 & P02\\
T51 \,\textperiodcentered\, \texttt{collect\_\allowbreak{}dark\_\allowbreak{}oak\_\allowbreak{}log} & dark oak log $\geq$1 & 450 & P02\\
T52 \,\textperiodcentered\, \texttt{craft\_\allowbreak{}dark\_\allowbreak{}oak\_\allowbreak{}planks} & dark oak planks $\geq$4 & 600 & P02\\
T53 \,\textperiodcentered\, \texttt{craft\_\allowbreak{}dark\_\allowbreak{}oak\_\allowbreak{}boat} & dark oak boat $\geq$1 & 700 & P02\\
T54 \,\textperiodcentered\, \texttt{craft\_\allowbreak{}bowl} & bowl $\geq$1 & 400 & P02\\
T55 \,\textperiodcentered\, \texttt{craft\_\allowbreak{}ladder} & ladder $\geq$1 & 500 & P02\\
T56 \,\textperiodcentered\, \texttt{craft\_\allowbreak{}oak\_\allowbreak{}sign} & oak sign $\geq$1 & 600 & P02\\
T57 \,\textperiodcentered\, \texttt{craft\_\allowbreak{}oak\_\allowbreak{}fence} & oak fence $\geq$1 & 600 & P02\\
T58 \,\textperiodcentered\, \texttt{craft\_\allowbreak{}oak\_\allowbreak{}door} & oak door $\geq$1 & 600 & P02\\
T59 \,\textperiodcentered\, \texttt{craft\_\allowbreak{}oak\_\allowbreak{}trapdoor} & oak trapdoor $\geq$1 & 600 & P02\\
T60 \,\textperiodcentered\, \texttt{craft\_\allowbreak{}oak\_\allowbreak{}pressure\_\allowbreak{}plate} & oak pressure plate $\geq$1 & 500 & P02\\
T61 \,\textperiodcentered\, \texttt{craft\_\allowbreak{}stone\_\allowbreak{}axe} & stone axe $\geq$1 & 800 & P02\\
T62 \,\textperiodcentered\, \texttt{craft\_\allowbreak{}stone\_\allowbreak{}shovel} & stone shovel $\geq$1 & 700 & P02\\
T63 \,\textperiodcentered\, \texttt{craft\_\allowbreak{}stone\_\allowbreak{}hoe} & stone hoe $\geq$1 & 700 & P02\\
T64 \,\textperiodcentered\, \texttt{smelt\_\allowbreak{}stone} & stone $\geq$1 & 900 & P02\\
\end{longtable}}
\smallskip
\Needspace{5\baselineskip}
\refstepcounter{apptable}\label{tab:app-f3}
{\small Table \theapptable. Supplementary Diversity and Exploration: 16 task definitions.\par}\nopagebreak\smallskip
{\fontsize{9}{10.7}\selectfont\setlength{\tabcolsep}{2pt}\renewcommand{\arraystretch}{1.10}
\begin{longtable}{>{\raggedright\arraybackslash}p{159.11pt}>{\raggedright\arraybackslash}p{143.96pt}>{\raggedright\arraybackslash}p{30.31pt}>{\raggedright\arraybackslash}p{45.46pt}}
\toprule
\textbf{Code / catalog task} & \textbf{Success condition} & \textbf{Steps} & \textbf{Profile}\\\midrule
\endfirsthead
\toprule
\textbf{Code / catalog task} & \textbf{Success condition} & \textbf{Steps} & \textbf{Profile}\\\midrule
\endhead
\bottomrule\endfoot
T65 \,\textperiodcentered\, \texttt{collect\_\allowbreak{}dirt} & dirt $\geq$1 & 200 & P02\\
T66 \,\textperiodcentered\, \texttt{collect\_\allowbreak{}sand} & sand $\geq$1 & 300 & P02\\
T67 \,\textperiodcentered\, \texttt{collect\_\allowbreak{}gravel} & gravel $\geq$1 & 350 & P02\\
T68 \,\textperiodcentered\, \texttt{collect\_\allowbreak{}flint} & flint $\geq$1 & 600 & P02\\
T69 \,\textperiodcentered\, \texttt{collect\_\allowbreak{}seeds} & wheat seeds $\geq$1 & 300 & P02\\
T70 \,\textperiodcentered\, \texttt{collect\_\allowbreak{}flower} & poppy $\geq$1 & 300 & P02\\
T71 \,\textperiodcentered\, \texttt{collect\_\allowbreak{}clay\_\allowbreak{}ball} & clay ball $\geq$1 & 400 & P02\\
T72 \,\textperiodcentered\, \texttt{collect\_\allowbreak{}cactus} & cactus $\geq$1 & 500 & P02\\
T73 \,\textperiodcentered\, \texttt{collect\_\allowbreak{}sugar\_\allowbreak{}cane} & sugar cane $\geq$1 & 500 & P02\\
T74 \,\textperiodcentered\, \texttt{collect\_\allowbreak{}apple} & apple $\geq$1 & 700 & P02\\
T75 \,\textperiodcentered\, \texttt{collect\_\allowbreak{}pumpkin} & pumpkin $\geq$1 & 700 & P02\\
T76 \,\textperiodcentered\, \texttt{collect\_\allowbreak{}melon\_\allowbreak{}slice} & melon slice $\geq$1 & 800 & P02\\
T77 \,\textperiodcentered\, \texttt{collect\_\allowbreak{}red\_\allowbreak{}mushroom} & red mushroom $\geq$1 & 700 & P02\\
T78 \,\textperiodcentered\, \texttt{collect\_\allowbreak{}brown\_\allowbreak{}mushroom} & brown mushroom $\geq$1 & 700 & P02\\
T79 \,\textperiodcentered\, \texttt{collect\_\allowbreak{}bone} & bone $\geq$1 & 900 & P02\\
T80 \,\textperiodcentered\, \texttt{collect\_\allowbreak{}emerald} & emerald $\geq$1 & 2200 & P12\\
\end{longtable}}
\smallskip
\Needspace{5\baselineskip}
\subsection{Ordered chain registry}
\label{app:suite-2}
A sequence below gives every task occurrence, including repetitions. Task codes expand through Tables~\ref{tab:app-f1}--\ref{tab:app-f3}. Each chain begins with the symbolic state \texttt{environment:spawned\_\allowbreak{}world}; no physical resource is created by that symbolic entry. The \texttt{night\_\allowbreak{}combat\_\allowbreak{}probe} chain additionally declares symbolic initial states \texttt{structure:lit\_\allowbreak{}shelter} and \texttt{tool:stone\_\allowbreak{}sword}; these declarations do not inject physical resources. ``Task'' selects the stage's task profile, whereas ``chain'' retains the listed chain profile. The runtime state-continuity rule is defined in Appendix~\ref{app:environment-4}.\par
\Needspace{5\baselineskip}
\refstepcounter{apptable}\label{tab:app-f4}
{\small Table \theapptable. Core Progression: exact chain sequences.\par}\nopagebreak\smallskip
{\fontsize{9}{10.7}\selectfont\setlength{\tabcolsep}{2pt}\renewcommand{\arraystretch}{1.10}
\noindent\begin{minipage}{\linewidth}\begin{tabular}{>{\raggedright\arraybackslash}p{138.15pt}>{\raggedright\arraybackslash}p{34.00pt}>{\raggedright\arraybackslash}p{211.60pt}}
\toprule
\textbf{Chain / profile / policy} & \textbf{Length} & \textbf{Ordered task codes}\\\midrule
\texttt{cross\_\allowbreak{}domain\_\allowbreak{}safe\_\allowbreak{}night}; P01; task & 10 & T01, T02, T03, T13, T04, T05, T06, T07, T08, T09\\
\texttt{combat\_\allowbreak{}readiness}; P01; task & 8 & T02, T03, T13, T04, T05, T10, T11, T12\\
\texttt{ore\_\allowbreak{}mining\_\allowbreak{}progression}; P12; chain & 19 & T02, T03, T13, T04, T05, T10, T23, T14, T01, T06, T15, T16, T27, T18, T19, T20, T21, T22, T80\\
\texttt{smelting\_\allowbreak{}transformations}; P12; chain & 21 & T02, T03, T13, T04, T05, T10, T14, T01, T06, T66, T02, T24, T10, T64, T25, T15, T16, T27, T19, T06, T26\\
\texttt{full\_\allowbreak{}mixed\_\allowbreak{}progression}; P01; task & 13 & T01, T02, T03, T13, T04, T05, T06, T07, T08, T09, T10, T11, T12\\
\bottomrule\end{tabular}\end{minipage}\par}
\smallskip
\Needspace{5\baselineskip}
\refstepcounter{apptable}\label{tab:app-f5}
{\small Table \theapptable. Transfer and Schema Generalization: exact chain sequences.\par}\nopagebreak\smallskip
{\fontsize{9}{10.7}\selectfont\setlength{\tabcolsep}{2pt}\renewcommand{\arraystretch}{1.10}
\noindent\begin{minipage}{\linewidth}\begin{tabular}{>{\raggedright\arraybackslash}p{138.15pt}>{\raggedright\arraybackslash}p{34.00pt}>{\raggedright\arraybackslash}p{211.60pt}}
\toprule
\textbf{Chain / profile / policy} & \textbf{Length} & \textbf{Ordered task codes}\\\midrule
\texttt{wood\_\allowbreak{}species\_\allowbreak{}crafting}; P02; task & 15 & T39, T40, T41, T42, T43, T44, T45, T46, T47, T48, T49, T50, T51, T52, T53\\
\texttt{oak\_\allowbreak{}utility\_\allowbreak{}crafting}; P02; task & 11 & T02, T03, T13, T04, T54, T55, T56, T57, T58, T59, T60\\
\texttt{stone\_\allowbreak{}tool\_\allowbreak{}suite}; P02; task & 10 & T02, T03, T13, T04, T05, T10, T23, T61, T62, T63\\
\texttt{wooden\_\allowbreak{}utility\_\allowbreak{}tool\_\allowbreak{}suite}; P02; task & 14 & T02, T03, T34, T02, T03, T35, T02, T03, T13, T04, T36, T02, T03, T38\\
\bottomrule\end{tabular}\end{minipage}\par}
\smallskip
\Needspace{5\baselineskip}
\refstepcounter{apptable}\label{tab:app-f6}
{\small Table \theapptable. Supplementary Diversity and Exploration: exact chain sequences.\par}\nopagebreak\smallskip
{\fontsize{9}{10.7}\selectfont\setlength{\tabcolsep}{2pt}\renewcommand{\arraystretch}{1.10}
\noindent\begin{minipage}{\linewidth}\begin{tabular}{>{\raggedright\arraybackslash}p{138.15pt}>{\raggedright\arraybackslash}p{34.00pt}>{\raggedright\arraybackslash}p{211.60pt}}
\toprule
\textbf{Chain / profile / policy} & \textbf{Length} & \textbf{Ordered task codes}\\\midrule
\texttt{basic\_\allowbreak{}resource\_\allowbreak{}sampler}; P02; task & 6 & T65, T66, T67, T68, T69, T70\\
\texttt{wood\_\allowbreak{}species\_\allowbreak{}collection}; P02; task & 5 & T39, T42, T45, T48, T51\\
\texttt{natural\_\allowbreak{}material\_\allowbreak{}sampler}; P02; task & 8 & T71, T72, T73, T74, T75, T76, T77, T78\\
\bottomrule\end{tabular}\end{minipage}\par}
\smallskip
\Needspace{5\baselineskip}
\refstepcounter{apptable}\label{tab:app-f7}
{\small Table \theapptable. Calibration and Probe: exact chain sequences.\par}\nopagebreak\smallskip
{\fontsize{9}{10.7}\selectfont\setlength{\tabcolsep}{2pt}\renewcommand{\arraystretch}{1.10}
\noindent\begin{minipage}{\linewidth}\begin{tabular}{>{\raggedright\arraybackslash}p{138.15pt}>{\raggedright\arraybackslash}p{34.00pt}>{\raggedright\arraybackslash}p{211.60pt}}
\toprule
\textbf{Chain / profile / policy} & \textbf{Length} & \textbf{Ordered task codes}\\\midrule
\texttt{collect\_\allowbreak{}log\_\allowbreak{}smoke}; P02; task & 1 & T02\\
\texttt{wooden\_\allowbreak{}tools\_\allowbreak{}smoke}; P02; task & 5 & T02, T03, T13, T04, T05\\
\texttt{coal\_\allowbreak{}smoke}; P05; chain & 7 & T02, T03, T13, T04, T05, T01, T06\\
\texttt{wood\_\allowbreak{}crafting\_\allowbreak{}basics}; P02; task & 5 & T02, T03, T13, T04, T05\\
\texttt{stone\_\allowbreak{}crafting\_\allowbreak{}basics}; P02; task & 7 & T02, T03, T13, T04, T05, T10, T14\\
\texttt{survival\_\allowbreak{}utility\_\allowbreak{}suite}; P01; task & 24 & T02, T03, T13, T04, T37, T17, T30, T79, T28, T02, T03, T29, T05, T10, T14, T01, T06, T15, T16, T02, T03, T31, T32, T33\\
\texttt{night\_\allowbreak{}combat\_\allowbreak{}probe}; P03; task & 2 & T09, T12\\
\bottomrule\end{tabular}\end{minipage}\par}
\smallskip
\clearpage
\subsection{Detailed direct capability results}
\label{app:direct-results}
The direct capability matrix in Table~\ref{tab:direct-capability} reports partition-level averages. Table~\ref{tab:app-direct-detail} expands those values to the 12 primary chains and the nine direct configurations. Each chain occupies three rows that report TSR, WGP, and VMP. The seven calibration and probe chains are retained in the complete chain registry but are not included in this expansion.
Table~\ref{tab:app-direct-protocol} fixes the protocol under which these direct results were collected.\par
{\parindent=0pt
\refstepcounter{apptable}\label{tab:app-direct-detail}
{\small Table \theapptable. Per-chain direct capability results for the 12 primary evaluation chains underlying Table~\ref{tab:direct-capability}. Each chain occupies three rows reporting TSR, WGP, and VMP in percent. The two planner-backed columns report GPT-5.6-sol runs. The nine scored direct configurations exclude the separately marked ExpeL runs. A superscript $2$ marks a two-seed report.\par}\nopagebreak\smallskip
\newcommand{\apphead}[1]{\begin{tabular}[t]{@{}c@{}}#1\end{tabular}}
\scriptsize
\setlength{\tabcolsep}{1.8pt}
\renewcommand{\arraystretch}{0.92}
\setlength{\aboverulesep}{1pt}
\setlength{\belowrulesep}{1pt}
\begin{tabular}{@{}lc*{9}{c}@{}}
\toprule
\rule{0pt}{2.2ex}\textbf{Chain} & \textbf{Metric} & \apphead{\textbf{STEVE-1}\\\textcolor{gray}{controller}} &
\apphead{\textbf{JARVIS-1}\\\textcolor{gray}{controller}} &
\apphead{\textbf{Optimus-1}\\\textcolor{gray}{controller}} &
\apphead{\textbf{Optimus-3}\\\textcolor{gray}{action path}} &
\apphead{\textbf{VPT}\\\textcolor{gray}{STEVE-1}\\\textcolor{gray}{ckpt.}} &
\apphead{\textbf{VPT}\\\textcolor{gray}{official 2x}\\\textcolor{gray}{foundation}} &
\apphead{\textbf{no-op}\\\textcolor{gray}{baseline}} &
\apphead{\textbf{JARVIS-1}\\\textcolor{gray}{planner}\\\textcolor{gray}{\raisebox{-0.15ex}{\includegraphics[height=1.0em]{fig/GPT.png}}}} &
\apphead{\textbf{Optimus-1}\\\textcolor{gray}{planner}\\\textcolor{gray}{\raisebox{-0.15ex}{\includegraphics[height=1.0em]{fig/GPT.png}}}}\\
\midrule
\rowcolor{black!10}\multicolumn{11}{l}{\textit{Core progression}}\\
\multirow{3}{*}{\shortstack[l]{\texttt{cross\_domain}\\\texttt{safe\_night}}} & \textcolor{gray}{TSR} & 30.0 & 16.7 & 26.7 & 10.0 & 13.3 & 10.0\rlap{\textsuperscript{2}} & 10.0 & 10.0 & 23.3\\
 & \textcolor{gray}{WGP} & 21.7 & 10.1 & 17.4 & 4.3 & 5.8 & 4.3 & 4.3 & 4.3 & 15.9\\
 & \textcolor{gray}{VMP} & 66.7 & 22.2 & 61.1 & 0.0 & 16.7 & 0.0 & 0.0 & 0.0 & 41.7\\
\arrayrulecolor{black!25}\midrule[0.3pt]\arrayrulecolor{black}
\multirow{3}{*}{\shortstack[l]{\texttt{combat}\\\texttt{readiness}}} & \textcolor{gray}{TSR} & 12.5 & 16.7 & 12.5 & 0.0 & 0.0 & 4.2 & 0.0 & 0.0 & 0.0\\
 & \textcolor{gray}{WGP} & 8.1 & 9.8 & 8.1 & 0.0 & 0.0 & 1.6 & 0.0 & 0.0 & 0.0\\
 & \textcolor{gray}{VMP} & 38.9 & 55.6 & 38.9 & 0.0 & 0.0 & 16.7 & 0.0 & 0.0 & 0.0\\
\arrayrulecolor{black!25}\midrule[0.3pt]\arrayrulecolor{black}
\multirow{3}{*}{\shortstack[l]{\texttt{full}\\\texttt{mixed\_progression}}} & \textcolor{gray}{TSR} & 20.5 & 12.8 & 23.1 & 7.7 & 7.7\rlap{\textsuperscript{2}} & 10.3 & 7.7 & 7.7 & 10.3\\
 & \textcolor{gray}{WGP} & 13.8 & 5.7 & 17.2 & 3.4 & 3.4 & 4.6 & 3.4 & 3.4 & 4.6\\
 & \textcolor{gray}{VMP} & 58.3 & 33.3 & 66.7 & 0.0 & 0.0 & 16.7 & 0.0 & 0.0 & 16.7\\
\arrayrulecolor{black!25}\midrule[0.3pt]\arrayrulecolor{black}
\multirow{3}{*}{\shortstack[l]{\texttt{ore}\\\texttt{mining\_progression}}} & \textcolor{gray}{TSR} & 0.0 & 3.5 & 0.0 & 0.0 & 0.0 & 0.0\rlap{\textsuperscript{2}} & 0.0\rlap{\textsuperscript{2}} & 0.0 & 0.0\\
 & \textcolor{gray}{WGP} & 0.0 & 2.9 & 0.0 & 0.0 & 0.0 & 0.0 & 0.0 & 0.0 & 0.0\\
 & \textcolor{gray}{VMP} & 0.0 & 22.2 & 0.0 & 0.0 & 0.0 & 0.0 & 0.0 & 0.0 & 0.0\\
\arrayrulecolor{black!25}\midrule[0.3pt]\arrayrulecolor{black}
\multirow{3}{*}{\shortstack[l]{\texttt{smelting}\\\texttt{transformations}}} & \textcolor{gray}{TSR} & 1.6 & 0.0 & 0.0 & 0.0 & 0.0 & 0.0 & 0.0 & 0.0 & 0.0\\
 & \textcolor{gray}{WGP} & 0.6 & 0.0 & 0.0 & 0.0 & 0.0 & 0.0 & 0.0 & 0.0 & 0.0\\
 & \textcolor{gray}{VMP} & 16.7 & 0.0 & 0.0 & 0.0 & 0.0 & 0.0 & 0.0 & 0.0 & 0.0\\
\midrule
\rowcolor{black!10}\multicolumn{11}{l}{\textit{Catalog transfer}}\\
\multirow{3}{*}{\shortstack[l]{\texttt{wood\_species}\\\texttt{crafting}}} & \textcolor{gray}{TSR} & 0.0 & 4.4 & 0.0 & 0.0 & 0.0 & 0.0 & 0.0 & 0.0 & 0.0\\
 & \textcolor{gray}{WGP} & 0.0 & 4.2 & 0.0 & 0.0 & 0.0 & 0.0 & 0.0 & 0.0 & 0.0\\
 & \textcolor{gray}{VMP} & 0.0 & 22.2 & 0.0 & 0.0 & 0.0 & 0.0 & 0.0 & 0.0 & 0.0\\
\arrayrulecolor{black!25}\midrule[0.3pt]\arrayrulecolor{black}
\multirow{3}{*}{\shortstack[l]{\texttt{oak\_utility}\\\texttt{crafting}}} & \textcolor{gray}{TSR} & 6.1 & 6.1 & 6.1 & 0.0 & 6.1 & 0.0 & 0.0 & 0.0 & 0.0\\
 & \textcolor{gray}{WGP} & 7.6 & 7.6 & 7.6 & 0.0 & 7.6 & 0.0 & 0.0 & 0.0 & 0.0\\
 & \textcolor{gray}{VMP} & 22.2 & 22.2 & 22.2 & 0.0 & 22.2 & 0.0 & 0.0 & 0.0 & 0.0\\
\arrayrulecolor{black!25}\midrule[0.3pt]\arrayrulecolor{black}
\multirow{3}{*}{\shortstack[l]{\texttt{stone\_tool}\\\texttt{suite}}} & \textcolor{gray}{TSR} & 16.7 & 10.0 & 13.3 & 0.0 & 0.0 & 6.7 & 0.0 & 0.0 & 6.7\\
 & \textcolor{gray}{WGP} & 14.4 & 6.5 & 7.8 & 0.0 & 0.0 & 2.6 & 0.0 & 0.0 & 5.2\\
 & \textcolor{gray}{VMP} & 47.2 & 38.9 & 55.6 & 0.0 & 0.0 & 33.3 & 0.0 & 0.0 & 22.2\\
\arrayrulecolor{black!25}\midrule[0.3pt]\arrayrulecolor{black}
\multirow{3}{*}{\shortstack[l]{\texttt{wooden\_utility}\\\texttt{tool\_suite}}} & \textcolor{gray}{TSR} & 7.1 & 14.3 & 9.5 & 0.0 & 0.0 & 2.4 & 0.0 & 0.0 & 0.0\rlap{\textsuperscript{2}}\\
 & \textcolor{gray}{WGP} & 5.6 & 13.6 & 9.0 & 0.0 & 0.0 & 1.1 & 0.0 & 0.0 & 0.0\\
 & \textcolor{gray}{VMP} & 38.9 & 66.7 & 44.4 & 0.0 & 0.0 & 16.7 & 0.0 & 0.0 & 0.0\\
\midrule
\rowcolor{black!10}\multicolumn{11}{l}{\textit{Diversity}}\\
\multirow{3}{*}{\shortstack[l]{\texttt{basic}\\\texttt{resource\_sampler}}} & \textcolor{gray}{TSR} & 5.6 & 5.6 & 5.6 & 0.0 & 0.0 & 5.6 & 0.0 & 0.0 & 0.0\\
 & \textcolor{gray}{WGP} & 5.6 & 5.6 & 5.6 & 0.0 & 0.0 & 5.6 & 0.0 & 0.0 & 0.0\\
 & \textcolor{gray}{VMP} & 16.7 & 16.7 & 16.7 & 0.0 & 0.0 & 16.7 & 0.0 & 0.0 & 0.0\\
\arrayrulecolor{black!25}\midrule[0.3pt]\arrayrulecolor{black}
\multirow{3}{*}{\shortstack[l]{\texttt{wood}\\\texttt{species\_collection}}} & \textcolor{gray}{TSR} & 0.0 & 0.0 & 0.0 & 0.0 & 0.0 & 10.0\rlap{\textsuperscript{2}} & 0.0 & 0.0 & 0.0\\
 & \textcolor{gray}{WGP} & 0.0 & 0.0 & 0.0 & 0.0 & 0.0 & 10.0 & 0.0 & 0.0 & 0.0\\
 & \textcolor{gray}{VMP} & 0.0 & 0.0 & 0.0 & 0.0 & 0.0 & 25.0 & 0.0 & 0.0 & 0.0\\
\arrayrulecolor{black!25}\midrule[0.3pt]\arrayrulecolor{black}
\multirow{3}{*}{\shortstack[l]{\texttt{natural}\\\texttt{material\_sampler}}} & \textcolor{gray}{TSR} & 0.0 & 0.0 & 0.0 & 0.0 & 0.0 & 0.0 & 0.0 & 0.0\rlap{\textsuperscript{2}} & 0.0\\
 & \textcolor{gray}{WGP} & 0.0 & 0.0 & 0.0 & 0.0 & 0.0 & 0.0 & 0.0 & 0.0 & 0.0\\
 & \textcolor{gray}{VMP} & 0.0 & 0.0 & 0.0 & 0.0 & 0.0 & 0.0 & 0.0 & 0.0 & 0.0\\
\bottomrule
\end{tabular}
\par}
\Needspace{10\baselineskip}
\refstepcounter{apptable}\label{tab:app-direct-protocol}
{\small Table \theapptable. Direct capability protocol. These runs measure immediate capability and do not define a source--target transfer relation.\par}\nopagebreak\smallskip
{\fontsize{9}{10.7}\selectfont\setlength{\tabcolsep}{2pt}\renewcommand{\arraystretch}{1.10}
\noindent\begin{minipage}{\linewidth}\begin{tabular}{>{\raggedright\arraybackslash}p{95pt}>{\raggedright\arraybackslash}p{\dimexpr\linewidth-103pt\relax}}
\toprule
\textbf{Item} & \textbf{Fixed setting}\\\midrule
Protocol identifier & \texttt{mcbench\_\allowbreak{}full19\_\allowbreak{}direct\_\allowbreak{}matrix\_\allowbreak{}v1}\\
Chains & All 19 registered chains are run; the 12 primary chains (5 core progression, 4 catalog transfer, 3 diversity) are scored, and the 7 calibration and probe chains are reported separately\\
Scale & Nine direct configurations $\times$ three seeds $\times$ 19 chains: 513 planned units, 505 valid reports; units without a valid report leave some cells with two seeds\\
Seeds & World seeds 3101, 3102, 3103; agent seeds 4101, 4102, 4103\\
Environment profiles & \texttt{per\_\allowbreak{}stage\_\allowbreak{}profiles=true}; each stage is initialized from its published profile\\
Observation and action & MineRL POV observation with a unified task description and action interface; the task graph, verifier predicates, and hidden resource coordinates are not exposed\\
Planner-backed runs & \texttt{gpt-\allowbreak{}5.6-\allowbreak{}sol} through the Responses interface with high reasoning effort; the planner produces subgoals only, and actions come from the underlying controller\\
Experience state & No source--target transfer; planner-backed configurations start from an empty, read-only memory\\
Reported fields & Stage records, TSR, WGP, VMP, environment steps, and termination status\\
ExpeL & One separately marked reference report per chain, excluded from the nine-configuration aggregates\\
\bottomrule\end{tabular}\end{minipage}\par}
\smallskip
Table~\ref{tab:app-stage-audit} gives the stage counts behind Figure~\ref{fig:stage-seed}a.\par
\Needspace{8\baselineskip}
\refstepcounter{apptable}\label{tab:app-stage-audit}
{\small Table \theapptable. Failure-stage audit over the 57 direct-matrix reports per configuration: passed / reached reports and conditional pass rate (\%) at the wood and crafting stages; the full-chain column counts reports whose entire chain passed.\par}\nopagebreak\smallskip
{\fontsize{9}{10.7}\selectfont\renewcommand{\arraystretch}{1.10}
\noindent\begin{tabular*}{\linewidth}{@{\extracolsep{\fill}}lccccc@{}}
\toprule
\textbf{Configuration} & \textbf{Collect log} & \textbf{Craft planks} & \textbf{Craft sticks} & \textbf{Crafting table} & \textbf{Full chain}\\\midrule
STEVE-1 & 24/42 (57.1) & 14/21 (66.7) & 2/13 (15.4) & 0/2 (0.0) & 3/57 (5.3)\\
JARVIS-1 (controller-only) & 22/42 (52.4) & 11/20 (55.0) & 1/8 (12.5) & 0/1 (0.0) & 2/57 (3.5)\\
Optimus-1 (controller-only) & 24/42 (57.1) & 16/23 (69.6) & 1/14 (7.1) & 0/1 (0.0) & 1/57 (1.8)\\
\bottomrule\end{tabular*}\par}
\smallskip
\clearpage
\Needspace{5\baselineskip}
\subsection{Source and target sequences for paired transfer}
\label{app:suite-3}
Table~\ref{tab:app-f8} expands the three transfer probe families into their exact source and target streams. Task codes refer to Tables~\ref{tab:app-f1}--\ref{tab:app-f3}. The shuffled condition reverses the related-source sequence, preserving its task multiset. The empty-artifact baseline has no source stream and uses the same target sequence.\par
\Needspace{5\baselineskip}
\refstepcounter{apptable}\label{tab:app-f8}
{\small Table \theapptable. Frozen source/target sequences. Repeated codes denote repeated stage occurrences.\par}\nopagebreak\smallskip
{\fontsize{9}{10.7}\selectfont\setlength{\tabcolsep}{2pt}\renewcommand{\arraystretch}{1.10}
\noindent\begin{minipage}{\linewidth}\begin{tabular}{>{\raggedright\arraybackslash}p{128.26pt}>{\raggedright\arraybackslash}p{260.42pt}}
\toprule
\textbf{Family / condition} & \textbf{Ordered task codes}\\\midrule
\texttt{wood\_\allowbreak{}species\_\allowbreak{}crafting} / Related & T39, T40\\
\texttt{wood\_\allowbreak{}species\_\allowbreak{}crafting} / Unrelated & T72, T73\\
\texttt{wood\_\allowbreak{}species\_\allowbreak{}crafting} / Shuffled & T40, T39\\
\texttt{wood\_\allowbreak{}species\_\allowbreak{}crafting} / Target & T48, T49\\
\texttt{wooden\_\allowbreak{}tool\_\allowbreak{}transfer} / Related & T02, T03, T13, T04, T36\\
\texttt{wooden\_\allowbreak{}tool\_\allowbreak{}transfer} / Unrelated & T71, T72, T73, T77, T70\\
\texttt{wooden\_\allowbreak{}tool\_\allowbreak{}transfer} / Shuffled & T36, T04, T13, T03, T02\\
\texttt{wooden\_\allowbreak{}tool\_\allowbreak{}transfer} / Target & T02, T03, T13, T04, T38\\
\texttt{stone\_\allowbreak{}tool\_\allowbreak{}transfer} / Related & T02, T03, T13, T04, T05, T10, T10, T23\\
\texttt{stone\_\allowbreak{}tool\_\allowbreak{}transfer} / Unrelated & T72, T73, T75, T76, T77, T78, T67, T71\\
\texttt{stone\_\allowbreak{}tool\_\allowbreak{}transfer} / Shuffled & T23, T10, T10, T05, T04, T13, T03, T02\\
\texttt{stone\_\allowbreak{}tool\_\allowbreak{}transfer} / Target & T02, T03, T13, T04, T05, T10, T10, T62\\
\bottomrule\end{tabular}\end{minipage}\par}
\smallskip
\clearpage
\section{Prompts and Reproducibility}
\label{app:reproduction}
\Needspace{5\baselineskip}
\subsection{Planning and reflection templates}
\label{app:reproduction-1}
The online-memory reference implementation uses two system templates. The planning request combines the current goal, an agent-visible state summary, and retrieved experience. It returns a small sequence of text subgoals for the visual controller. The reflection request summarizes a completed source episode for later retrieval. Cards~\ref{card:app-g1} and~\ref{card:app-g2} specify the input--output contracts and system prompts.\par
\refstepcounter{appcard}\label{card:app-g1}
\begin{tcolorbox}[title={Card \theappcard. Online-memory planning}]
\textbf{Purpose.} Translate the current goal and retrieved experience into controller subgoals.\par
\textbf{Input.} \texttt{goal}, \texttt{visible\_\allowbreak{}state}, \texttt{retrieved\_\allowbreak{}experience}\par
\textbf{Output.} JSON \texttt{steps} with 1--6 text prompts.\par
\begin{Verbatim}[fontsize=\fontsize{8.5}{10}\selectfont,numbers=left,numbersep=5pt,xleftmargin=13pt,frame=none,breaklines=true,breakanywhere=true]
You are a Minecraft task planner controlling a visual low-level policy. Use only the
natural-language goal, visible state, and prior experience. Do not assume access to reward
predicates, task graphs, hidden world coordinates, or tools outside Minecraft. Return JSON
only: {"steps":["short visual action prompt", ...]}. Provide 1 to 6 concise prompts
suitable for a STEVE-I-like action controller.
\end{Verbatim}
\end{tcolorbox}
\refstepcounter{appcard}\label{card:app-g2}
\begin{tcolorbox}[title={Card \theappcard. Experience reflection}]
\textbf{Purpose.} Convert source-stage feedback into reusable advice.\par
\textbf{Input.} Goal, pre-action state, dispatched prompts, stage feedback.\par
\textbf{Output.} One practical sentence stored as \texttt{advice}.\par
\begin{Verbatim}[fontsize=\fontsize{8.5}{10}\selectfont,numbers=left,numbersep=5pt,xleftmargin=13pt,frame=none,breaklines=true,breakanywhere=true]
You summarize one Minecraft episode for later reuse. Use only the supplied goal,
agent-visible state summary, controller dispatch evidence, and stage feedback. The plan is
proposed text, not proof that every prompt was attempted. Only dispatched_plan entries were
sent to the controller and returned actions; this does not prove their intended effects
occurred. The visible_state is the last pre-action observation, not a post-action terminal
observation. Do not invent hidden information or unobserved events. Return one concise
practical sentence.
\end{Verbatim}
\end{tcolorbox}
The planning payload contains \texttt{goal}, \texttt{visible\_\allowbreak{}state}, and \texttt{retrieved\_\allowbreak{}experience}. Retrieved records expose the instruction, plan, status, and advice. The planning parser returns up to six non-empty steps, using the task instruction as its default. Reflection receives stage feedback together with the actually dispatched prompts, allowing the stored advice to distinguish planned behavior from observed execution context.\par
\Needspace{5\baselineskip}
\subsection{Persistent memory representation}
\label{app:reproduction-2}
The artifact is a JSON object containing a record count and a list of episode records. Each record associates a task instruction with a proposed plan, the prompts dispatched to the controller, terminal status, interaction count, visible state, and advice. The memory writer keeps the latest configured number of records and replaces the artifact atomically. Retrieval and update rules are specified in Appendix~\ref{app:agents}; the schema below describes the fields required to inspect the stored experience.\par
\Needspace{5\baselineskip}
\refstepcounter{apptable}\label{tab:app-g1}
{\small Table \theapptable. Persistent memory fields and their interpretation.\par}\nopagebreak\smallskip
{\fontsize{9}{10.7}\selectfont\setlength{\tabcolsep}{2pt}\renewcommand{\arraystretch}{1.10}
\noindent\begin{minipage}{\linewidth}\begin{tabular}{>{\raggedright\arraybackslash}p{88.26pt}>{\raggedright\arraybackslash}p{134.32pt}>{\raggedright\arraybackslash}p{161.18pt}}
\toprule
\textbf{Field} & \textbf{Content} & \textbf{Role in reproduction}\\\midrule
\texttt{format} & Memory schema identifier & Identifies the memory format\\
\texttt{record\_\allowbreak{}count} & Number of retained records & Checks artifact size\\
\texttt{instruction} & Task text & Retrieval key\\
\texttt{plan} & Proposed controller prompts & Records planner output\\
\texttt{dispatched\_\allowbreak{}plan} & Prompts actually sent to the controller & Links planning to dispatch\\
\texttt{status}, \texttt{steps} & Stage outcome and executed steps & Captures feedback supplied to the method\\
\texttt{visible\_\allowbreak{}state} & Agent-side state summary & Records the reflection context\\
\texttt{advice} & Concise text retained for reuse & Reusable experience content\\
\bottomrule\end{tabular}\end{minipage}\par}
\smallskip
The source artifact is copied or otherwise identified before target evaluation, and its fingerprint accompanies the target role. Read-only target memory keeps the supplied experience fixed throughout evaluation. This makes the artifact itself an inspectable component of a paired comparison.\par
\Needspace{5\baselineskip}
\subsection{ExpeL action and rule templates}
\label{app:reproduction-3}
The ExpeL adaptation uses separate system templates for selecting an environment action and extracting reusable experience. Cards~\ref{card:app-g3} and~\ref{card:app-g4} specify the action and reflection interfaces. The action payload contains \texttt{instruction}, \texttt{observation}, \texttt{experiences}, and \texttt{insights}. The final field combines insight text with retrieved rules formatted as \texttt{[Expe\allowbreak{}L rule weight=...] ...}. This representation exposes the rule's persistence weight alongside its content. The observation is the structured state summary supplied by the adapter.\par
\refstepcounter{appcard}\label{card:app-g3}
\begin{tcolorbox}[title={Card \theappcard. ExpeL action selection}]
\textbf{Purpose.} Map visible state and retrieved experience to a game action.\par
\textbf{Input.} \texttt{instruction}, \texttt{observation}, \texttt{experiences}, \texttt{insights}\par
\textbf{Output.} A JSON \texttt{action} mapping for MineRL.\par
\begin{Verbatim}[fontsize=\fontsize{8.5}{10}\selectfont,numbers=left,numbersep=5pt,xleftmargin=13pt,frame=none,breaklines=true,breakanywhere=true]
You are the adapted ExpeL Minecraft policy for MC-EvoBench. Use only the task,
agent-visible observation, retrieved episodes, and insights. Never infer hidden verifier
predicates or coordinates. Return JSON only with an action object suitable for MineRL, e.g.
{"action": {"forward": 1}}. The observation may contain public nearby_blocks counts and
one-hot line_of_sight block indicators. If line_of_sight contains the requested resource
(for example oak_log=1), preserve the aim: when its numeric distance is at most 4, use
{"attack": 1} to mine it; when it is farther away, use {"forward": 1} without turning so
the agent can approach and collect the drop. Use small camera changes to search only when
the target is not in line of sight.
\end{Verbatim}
\end{tcolorbox}
The action parser accepts a mapping directly or JSON text, including a fenced JSON response. It accepts either a nested \texttt{action} mapping or the mapping itself. A singleton \texttt{\{"noop": true\}} or \texttt{\{"noop": 1\}} becomes the adapter's no-op marker. The action parser validates the response mapping; task success is evaluated from environment observations.\par
\refstepcounter{appcard}\label{card:app-g4}
\begin{tcolorbox}[title={Card \theappcard. ExpeL rule update}]
\textbf{Purpose.} Extract reusable experience and update persistent rules.\par
\textbf{Input.} Stage observations, dispatched actions, and status.\par
\textbf{Output.} JSON \texttt{insight} and \texttt{operations}.\par
\begin{Verbatim}[fontsize=\fontsize{8.5}{10}\selectfont,numbers=left,numbersep=5pt,xleftmargin=13pt,frame=none,breaklines=true,breakanywhere=true]
Extract one concise reusable insight from this Minecraft episode and update the persistent
ExpeL rule list. Use only supplied observations, dispatched actions, and status; do not
claim an effect unless it is explicitly observed. Return JSON only: {"insight": "...",
"operations": [{"op": "ADD|EDIT|REMOVE|AGREE", "rule_number": 1, "rule": "..."}]}. Use at
most four operations and use an empty operations list when no durable general rule is
justified.
\end{Verbatim}
\end{tcolorbox}
Reflection receives the stage episode: \texttt{instruction}, \texttt{status}, \texttt{steps}, \texttt{actions}, \texttt{visible\_\allowbreak{}state}, \texttt{visible\_\allowbreak{}state\_\allowbreak{}timing}, and \texttt{method}. The state is explicitly labeled as the last pre-action observation, and the action list includes each returned action, including repetitions. The reflection parser accepts \texttt{insight} or \texttt{reflection} for the insight, and \texttt{operations} or \texttt{rules} for the operation list. It normalizes the response into insight text and rule operations. The prompt requests up to four rule operations per reflection.\par
\Needspace{5\baselineskip}
\refstepcounter{apptable}\label{tab:app-g2}
{\small Table \theapptable. ExpeL persistent artifact fields. Episode records, extracted insights, and rule-update history are stored separately.\par}\nopagebreak\smallskip
{\fontsize{9}{10.7}\selectfont\setlength{\tabcolsep}{2pt}\renewcommand{\arraystretch}{1.10}
\noindent\begin{minipage}{\linewidth}\begin{tabular}{>{\raggedright\arraybackslash}p{128.26pt}>{\raggedright\arraybackslash}p{260.42pt}}
\toprule
\textbf{Field} & \textbf{Stored content}\\\midrule
\texttt{format} & ExpeL memory schema identifier\\
\texttt{record\_\allowbreak{}count}, \texttt{insight\_\allowbreak{}count} & Sizes of the retained episode and insight lists\\
\texttt{records} & Stage episodes with action and feedback fields\\
\texttt{insights} & Whitespace-normalized reusable text\\
\texttt{rules} & Rule objects containing \texttt{text} and integer \texttt{weight}\\
\texttt{rule\_\allowbreak{}events} & Normalized \texttt{operations}, \texttt{list\_\allowbreak{}full}, and the resulting \texttt{rules} snapshot\\
\bottomrule\end{tabular}\end{minipage}\par}
\smallskip
The artifact therefore preserves both the interaction record and the rule state used for subsequent decisions. The loader validates the artifact's format identifier.\par
\Needspace{5\baselineskip}
\subsection{Configuration example}
\label{app:reproduction-4}
Card~\ref{card:app-g5} gives the protocol fields from the wood-species target-after-related configuration. It includes the selected environment, agent randomness, memory mode, and target sequence. The complete repository file also specifies checkpoint locations, planner settings, artifact paths, role metadata, and output locations. Deployment paths and artifact format identifiers are resolved through \texttt{reproduction/\allowbreak{}paths.json}.\par
\refstepcounter{appcard}\label{card:app-g5}
\begin{tcolorbox}[title={Card \theappcard. Matched target configuration}]
\textbf{Purpose.} Fix the target conditions while selecting a source artifact.\par
\textbf{Input.} A frozen source artifact and the wood-species probe.\par
\textbf{Output.} A read-only target run under the listed field values.\par
\begin{Verbatim}[fontsize=\fontsize{8.5}{10}\selectfont,numbers=left,numbersep=5pt,xleftmargin=13pt,frame=none,breaklines=true,breakanywhere=true]
{
  "environment_profile": "evolution_wood_species_grove",
  "max_stage_steps": 450,
  "environment": {
    "kind": "minerl",
    "seed": 31,
    "per_stage_profiles": false
  },
  "agent": {
    "kind": "jarvis1_online",
    "random_seed": 5100,
    "memory_mode": "read_only",
    "memory_max_records": 256,
    "retrieval_limit": 4,
    "skill_horizon_steps": 120
  },
  "probe": {
    "profile_policy": "chain",
    "task_ids": [
      "collect_acacia_log",
      "craft_acacia_planks"
    ]
  }
}
\end{Verbatim}
\end{tcolorbox}
The target collects acacia logs and crafts acacia planks while using the artifact generated by the related birch source. The chain profile remains constant within this episode. Stage limits are applied by the runner, and task success uses the catalog predicates. The source role uses its own task sequence and enables memory writing; all target roles retain the same task definition.\par
\Needspace{5\baselineskip}
\subsection{Evaluation algorithm}
\label{app:reproduction-5}
The following pseudocode summarizes the common orchestration contract. Source interaction, artifact persistence, environment reset, and target evaluation are separate operations. The same control flow can host parameter updates, text memory, or executable skills, provided that the permitted persistent state and input interface are declared.\par
\refstepcounter{appcard}\label{card:app-g6}
\begin{tcolorbox}[title={Card \theappcard. Paired experience evaluation}]
\textbf{Purpose.} Isolate the effect of related source experience.\par
\textbf{Input.} Probe family, paired seed, base agent, matched source budget.\par
\textbf{Output.} Four matched target scores and within-seed contrasts.\par
\begin{Verbatim}[fontsize=\fontsize{8.5}{10}\selectfont,numbers=left,numbersep=5pt,xleftmargin=13pt,frame=none,breaklines=true,breakanywhere=true]
for each probe family and paired seed:
    base = load_base_agent_state()
    score[0] = evaluate_target(base, fresh_target_world())
    for condition in [related, unrelated, shuffled]:
        state = independent_copy(base)
        remaining = source_budget[family]
        while remaining > 0:
            episode = run_source_episode(state, condition, remaining)
            update_source_artifact(state, episode)
            remaining -= episode.environment_steps
        artifact = freeze_and_identify(state.persistent_state)
        target_agent = load_for_read_only_evaluation(base, artifact)
        score[condition] = evaluate_target(
            target_agent, fresh_target_world())
    save_paired_contrasts(score, source_costs, artifact_identities)
\end{Verbatim}
\end{tcolorbox}
At the task level, evaluation first observes the initial state, then alternates action and predicate evaluation until a terminal condition is reached. At the chain level, the runner applies the selected profile policy and records the first non-passing stage. At the experiment level, matched roles are combined within a seed before aggregation. These layers align environment execution, task success, and experience comparison at their respective units of analysis.\par
\Needspace{5\baselineskip}
\subsection{Reproduction entry points}
\label{app:reproduction-6}
Table~\ref{tab:app-g3} identifies the small set of source components that define the benchmark. The accompanying task and chain CSV files provide compact access to the catalog, while the JSON source files retain the executable definitions.\par
\Needspace{5\baselineskip}
\refstepcounter{apptable}\label{tab:app-g3}
{\small Table \theapptable. Source map for reproducing the definitions in this appendix.\par}\nopagebreak\smallskip
{\fontsize{9}{10.7}\selectfont\setlength{\tabcolsep}{2pt}\renewcommand{\arraystretch}{1.10}
\noindent\begin{minipage}{\linewidth}\begin{tabular}{>{\raggedright\arraybackslash}p{128.26pt}>{\raggedright\arraybackslash}p{260.42pt}}
\toprule
\textbf{Component} & \textbf{Repository location}\\\midrule
Tasks, states, edges, chains & \texttt{catalog} in \texttt{reproduction/\allowbreak{}paths.json}\\
Schema, graph, compilation & \texttt{src/\allowbreak{}mcbench/\allowbreak{}schema.py}, \texttt{graph.py}, \texttt{chains.py}\\
Environment and profiles & \texttt{src/\allowbreak{}mcbench/\allowbreak{}adapters/\allowbreak{}minerl.py}, \texttt{profiles.py}\\
Predicates and stage loop & \texttt{src/\allowbreak{}mcbench/\allowbreak{}verification.py}, \texttt{runner.py}\\
Execution and transfer metrics & \texttt{src/\allowbreak{}mcbench/\allowbreak{}metrics.py}, \texttt{evolution.py}\\
Paired transfer plan and profiles & \texttt{paired\_\allowbreak{}plan}, \texttt{paired\_\allowbreak{}profiles} in \texttt{reproduction/\allowbreak{}paths.json}\\
Reference memory implementation & \texttt{src/\allowbreak{}mcbench/\allowbreak{}adapters/\allowbreak{}jarvis1\_\allowbreak{}online.py}\\
Configuration execution & \texttt{src/\allowbreak{}mcbench/\allowbreak{}cli.py}\\
\bottomrule\end{tabular}\end{minipage}\par}
\smallskip
Catalog validation checks the task definitions and compiled chains. A dry run then exercises the reporting path with the repository's diagnostic adapter. Real execution additionally requires the selected runtime, policy checkpoints, and method-specific dependencies. The experiment configuration records these choices alongside the catalog snapshot. Card~\ref{card:app-g7} runs from the appendix bundle directory and resolves the catalog through the companion path map.\par
\refstepcounter{appcard}\label{card:app-g7}
\begin{tcolorbox}[title={Card \theappcard. Reproduction entry points}]
\textbf{Purpose.} Validate the catalog and inspect the reporting path.\par
\textbf{Input.} The companion path map and repository source.\par
\textbf{Output.} Catalog checks and a diagnostic chain report.\par
\begin{Verbatim}[fontsize=\fontsize{8.5}{10}\selectfont,numbers=left,numbersep=5pt,xleftmargin=13pt,frame=none,breaklines=true,breakanywhere=true]
python3 tools/reproduce.py validate
python3 tools/reproduce.py dry-run cross_domain_safe_night
\end{Verbatim}
\end{tcolorbox}
A reproducible experiment release associates the frozen catalog, configuration, model identity, seeds, and artifact identity with each chain report. Stage evidence and source-cost records preserve the path from a task definition to a paired contrast.\par